\documentclass[A4]{amsart}
\input xypic
\input xy \xyoption{all}

\let\originallhook\lhook
\let\originalrhook\rhook
\usepackage{MnSymbol}
\let\lhook\originallhook
\let\rhook\originalrhook

\usepackage{bbm}
\usepackage{amsmath}
\usepackage{algorithm} 
\usepackage{algorithmic}
\usepackage{graphicx}
\usepackage{caption}
\usepackage{subcaption}
\usepackage{epstopdf}
\usepackage{hyperref}

\theoremstyle{definition}

\newcounter{bean}

\newcommand{\seqm}[3]{\ensuremath{#1\stackrel{#2}
 {\longrightarrow}#3}}

\newcommand{\paren}[1]{\ensuremath{\left( #1 \right)}}
\newcommand{\br}[1]{\ensuremath{\left\{ #1 \right\}}}

\newcommand{\cunion}[3]{\displaystyle\bigcup^{#2}_{#1}#3}

\newcommand{\csum}[3]{\displaystyle\sum^{#2}_{#1}#3}

\newcommand{\cset}[2]{\br{#1\,\,\middle\vert\,\,#2}}

\newcommand{\mc}[1]{\ensuremath{\mathcal{#1}}}
\newcommand{\mb}[1]{\ensuremath{\mathbb{#1}}}

\newcommand{\bd}{\ensuremath{\partial}}

\newcommand{\wcolon}{\ensuremath{\,\colon\,}}

\begin{document}
\title{Fitting a Simplicial Complex Using a Variation of \MakeLowercase{k}-means}

\author{Piotr Beben}
\address{\scriptsize{School of Mathematics, University of Southampton, Southampton SO17 1BJ, United Kingdom}} 
\email{P.D.Beben@soton.ac.uk; pdbcas@gmail.com} 


\begin{abstract}

We give a simple and effective two stage algorithm for approximating a point cloud $\mc S\subset\mb R^m$ by a simplicial complex $K$.
The first stage is an iterative fitting procedure that generalizes k-means clustering, while the second stage involves deleting redundant simplices. 
A form of dimension reduction of $\mc S$ is obtained as a consequence. 

\end{abstract}

\maketitle

\section{Introduction}

The use of simplicial complexes as a means for estimating topology has found many applications to data analysis in recent times. 
For example, unsupervised learning techniques such as persistent homology~\cite{MR2476414,MR2405684} often use what are known as
Cech or Vietoris-Rips filtrations to capture multi-scale topological features of a point cloud $\mc S\subset\mb R^m$.
The simplicial complexes in these filtrations individually are not always a good representation of the 
actual physical shape of $\mc S$ since, for example, they often have a higher dimension than $\mb R^m$.
Our aim is to give an algorithm that can approximate $\mc S$ to the greatest extent possible when fed any simplicial complex $K$ mapped linearly into 
$\mb R^m$.
This algorithm has several nice properties, including a tendency towards preserving embeddings, 
as well as reducing to k-means clustering when $K$ is $0$-dimensional.
The resulting fitting is further refined by deleting simplices that have been poorly matched with $\mc S$;
the end result being a locally linear approximation of $\mc S$.
A lower dimensional representation of $\mc S$ in terms of barycentric coordinates follows by projecting onto this approximation.

\section{Algorithm Description}

Fix $K$ be a (geometric) simplicial complex, and let $\{v_1,\ldots,v_n\}\in K$ be the set of vertices of $K$.
The \emph{facets} of $K$ are the simplices of $K$ that have the highest dimension in the sense that they are not contained 
in the boundary of any other simplex. $K$ may then be represented as a collection of facets, 
each of which is represented by the set of vertices that it contains. 
The dimension of a facet is equal to the number of vertices it contains minus $1$. 
When we refer to the \emph{boundary} $\bd\sigma$ of a simplex $\sigma$, we mean the union of its smaller dimensional boundary simplices, 
even when the simplex is embedded as a subset of $\mb R^m$. Its \emph{interior} $int(\sigma)$ is $\sigma$ minus this boundary. 

Any point $x\in K$ is contained in a unique smallest dimensional simplex $\sigma_x\subseteq K$.
We may represent $x$ uniquely as a convex combination
$$
x=\csum{v_j\in\sigma_x}{}{\lambda_{jx} v_j}
$$
over the vertices $v_j$ that are contained in $\sigma_x$,  
where $\csum{v_j\in\sigma_x}{}{\lambda_{jx}}=1$ for some \emph{barycentric coordinates} $\lambda_{jx}\geq 0$. 
A map $g\colon\seqm{K}{}{\mb R}$ is said to be \emph{linear} if it is linear on each simplex of $K$.
Namely, 
$$
g(x)=\csum{v_j\in\sigma_x}{}{\lambda_{jx} g(v_j)}.
$$ 
So $g$ restricts to a linear embedding on each simplex of $K$, 
and is uniquely determined by the values $g(v_j)$ that it takes on its vertices $v_j$. 
Thus, we have a convenient representation of $g$ in terms of the $g(v_j)$'s.
 
\subsection{Fitting}

Fix $\mc S\subset\mb R^m$ our finite set of data points. 
Suppose $f\colon\seqm{K}{}{\mb R^m}$ is any choice of linear map, meant to represent our \emph{initial fitting} of $K$ to $S$. 
Starting with $f$, our aim is to obtain successively better fittings $f^{\ell}\colon\seqm{K}{}{\mb R^m}$ of $K$ to $\mc S$,
at each iteration giving a better reflection of the shape and structure of $\mc S$. We do this as follows.

\begin{algorithm}[H]
\caption{(Simplicial Means) First stage: fitting $K$ to $\mc S$}
\label{LA1}
\begin{algorithmic}[1]
\STATE set $f^{0} \leftarrow f$ and $\ell \leftarrow -1$;
\REPEAT 
\STATE increment $\ell \leftarrow \ell+1$;
\FOR{each $y\in \mc S$} 
\STATE find a choice of $y'\in K$ such that $f^\ell(y')$ is nearest to $y$,
and $\sigma_{y}\subseteq K$ the smallest dimensional simplex that contains $y'$;
\STATE compute $\lambda_{jy}\geq 0$ such that $\csum{v_j\in\sigma_y}{}{\lambda_{jy}}=1$ and $y'$ is the convex combination 
$$
y'=\csum{v_j\in\sigma_y}{}{\lambda_{jy} v_j}.
$$ 
\ENDFOR
\STATE\label{LSassign} using $f^\ell$, construct a new linear map 
$f^{\ell+1}\colon\seqm{K}{}{\mb R^m}$, defined on each vertex $v_j$ by setting 
$$
f^{\ell+1}(v_j) \leftarrow \frac{1}{|\mc N^\ell_j|}\csum{y\in\mc N^\ell_j}{}{\paren{(1-\lambda_{jy})f^{\ell}(v_j)+\lambda_{jy}y}}
$$
when $|\mc N^\ell_j|>0$, and $f^{\ell+1}(v_j)\leftarrow f^\ell(v_j)$ otherwise, where
$$
\mc N^\ell_j := \cset{y\in \mc S}{v_j\in\sigma_y}.
$$
\UNTIL{a given stop condition has not been reached 
(e.g. $\max_j ||f^{\ell+1}(v_j)-f^{\ell}(v_j)||$ is small)}
\STATE set $g \leftarrow f^\ell$;
\STATE \textbf{return} $g$, each $\sigma_{y}$, $y'$, and the barycentric coordinates $\lambda_{jy}$.
\end{algorithmic}
\end{algorithm}

\begin{figure}[H]
    \centering
    \begin{subfigure}[h]{0.3\textwidth}
        \includegraphics[width=\textwidth]{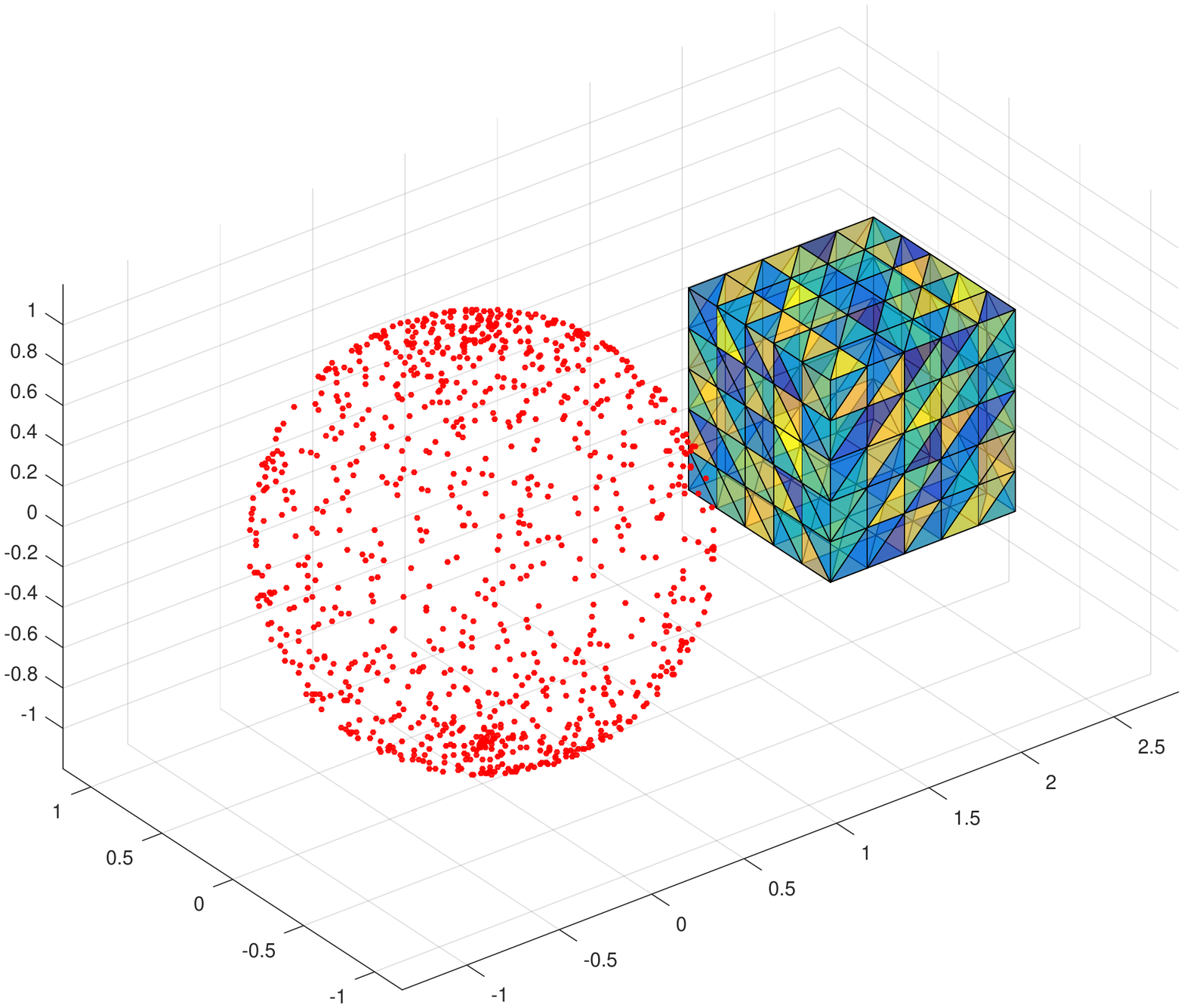}
        \caption{$0$ iterations}
    \end{subfigure}
    \begin{subfigure}[h]{0.3\textwidth}
        \includegraphics[width=\textwidth]{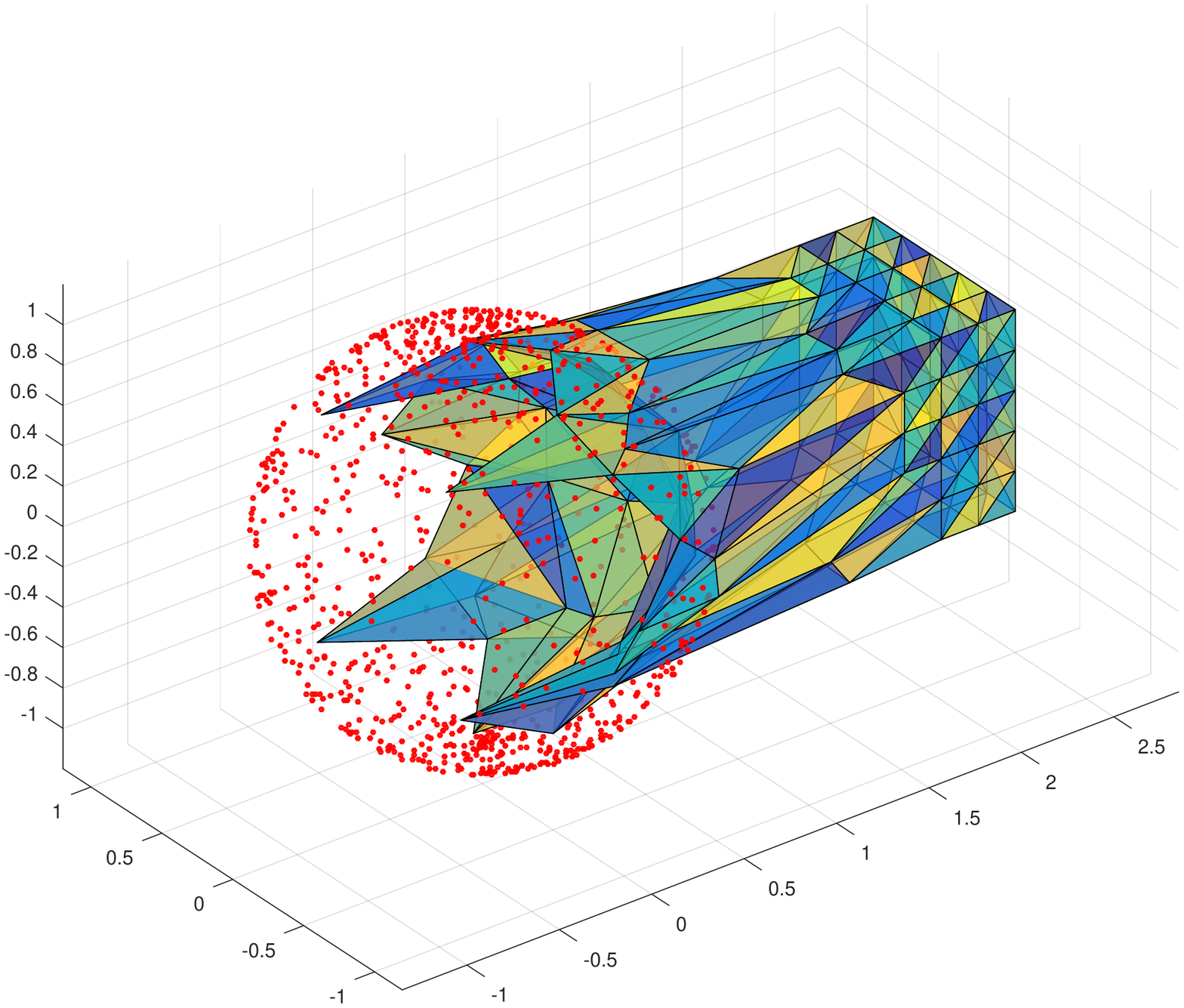}
        \caption{$10$ iterations}
    \end{subfigure}
    \begin{subfigure}[h]{0.3\textwidth}
        \includegraphics[width=\textwidth]{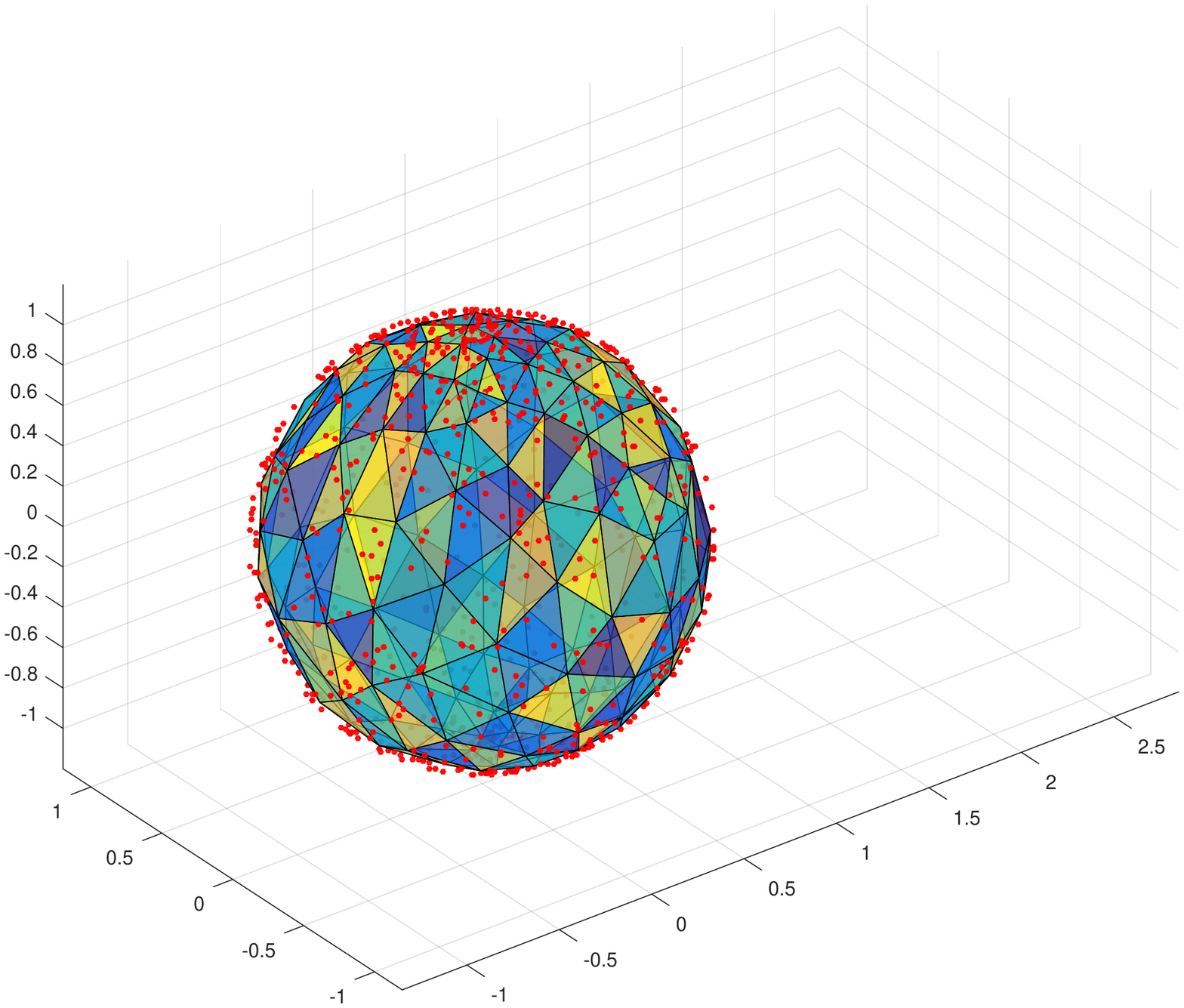}
        \caption{$150$ iterations}
    \end{subfigure}
    \caption{$1000$ points sampled from a unit $2$-sphere, $K$ the triangulated boundary of a $5\times 5\times 5$ mesh.
		Fitted using the sets $\overline{\mc N}^\ell_j$ in place of $\mc N^\ell_j$.}\label{LF8}
\end{figure}

The $|\mc N^\ell_j|>0$ assignment for $f^{\ell+1}(v_j)$ in Step~\ref{LSassign} can be generalized to 
$$
f^{\ell+1}(v_j) \leftarrow \frac{1}{|\mc N^\ell_j|}\csum{y\in\mc N^\ell_j}{}{\paren{\,\paren{\frac{1-\lambda_{jy}}{1+s}}f^\ell(v_j)+\paren{\frac{\lambda_{jy}+s}{1+s}}y\,}}
$$
where $s\geq 0$ is the \emph{learning rate}. Larger values for $s$ lead to faster convergence, but poorer fittings and reduced stability. 
Taking $s$ to be small ($\leq 0.1$) but nonzero has every advantage in addition to preventing the change in the mapping of a vertex $v_j$ 
becoming sluggish between iterations when the barycentric coordinates $\lambda_{jy}$ are all near zero. 
On the other hand, this does not prevent mapping of $v_j$ from becoming stuck in its current position when each $\lambda_{jy}$ is zero (regardless of the value $s$)
since $\mc N^\ell_j$ consists only of those $y\in\mc S$ for which $y'$ is on the interior of some simplex $\sigma$ that contains $v_j$
(meaning those on the boundary faces of $\sigma$ opposite to $v_j$ are ignored). There are advantages and disadvantages to this. 
On the positive side, higher dimensional simplices are often prevented from being collapsed onto lower dimensional linear patches of data,
but this can also be prevent simplices from being fitted in situations where it would be desired. 
For example, when fitting $K$ homeomorphic to a sphere to data sampled from a sphere (Figure~\ref{LF8}), 
\emph{craters} can form and remain in subsequent iterations if data points surrounding the crater are in fact nearest to points lying on its rim. 
This issue can be circumvented (but our advantages reversed!) by taking the slightly larger set 
$$
\overline{\mc N}^\ell_j=\cset{y\in \mc S}{y'\in\sigma\mbox{ for some simplex }\sigma\mbox{ such that }v_j\in\sigma}
$$
in place of $\mc N^\ell_j$, together with $s>0$.

There is a straightforward intuition underlying Algorithm~\ref{LA1}.
Consider the case where at the $\ell^{th}$ iteration $f^\ell$ is a linear embedding. 
Then $v_j$ and $K$ can then be thought of as $f^\ell(v_j)$ and the subspace $f^\ell(K)\subseteq \mb R^m$,
and Algorithm~\ref{LA1} in essence has $\mc S$ attract $K$ towards it by having each point $y\in \mc S$ exert a pull on a nearest point $y'\in K$.  
The caveat here is that in doing so the embedding $f^\ell$ must be kept linear,
so the net effect of this pull must come down to its influence on the individual vertices $v_j$ of the simplex containing $y'$.
The influence on each of these vertices $v_j$ should in turn decrease with some measure of distance of $y'$ to $v_j$.
This is analogous to pulling on a string attached at some point along a perpendicular uniform rod floating in space,
in which case the acceleration of an endpoint of the rod in the direction of the pull decreases with increasing distance from the string.
Since the size or shape of the simplex in our context is irrelevant, 
the distance from $y'$ to $v_j$ is measured in terms of its barycentric coordinates $\lambda_{jy}$.
In particular, if $y'$ lies near a boundary simplex opposite to $v_j$, then $\lambda_{jy}$ is near $0$ and $y$ has little influence on $v_j$,
while $y$ has full influence on $v_j$ when $\lambda_{jy}=1$, or equivalently, when $y'=v_j$.
The accumulation of these pulling forces on a vertex $v_j$ leads us to take the centroid of the $(1-\lambda_{jy})v_j+\lambda_{jy}y$
over all $y\in N^\ell_j$; equivalently, over all $y\in\mc S$ that are closest to a point $y'$ lying on the interior of a simplex that has $v_j$ as a vertex.   

When $K$ is $0$-dimensional -- namely a collection of $n$ disjoint vertices $v_j$ --
there is only a single barycentric coordinate $\lambda_{jy}=1$ for each $y'$ and $\mc N^\ell_j=\cset{y\in \mc S}{y\mbox{ nearest to }f^\ell(v_j)}$,
so Algorithm~\ref{LA1} reduces to the classical k-means algorithm with initial clusters $f^0(v_j)$.
Algorithm~\ref{LA1} is thereby a high dimensional non-discrete generalization of k-means clustering.
This is perhaps in the same spirit as persistent homology is a higher dimensional generalization of hierarchical clustering
(also by way of simplicial complexes).

\subsection{Preserving embeddings}

To obtain a our fitting $g$, why not simply apply the k-means algorithm to the vertices $f(v_j)$?
This would likely be a poor fitting of $K$ since the arrangement of simplices comprising $K$ is ignored completely.
Moreover, $g$ would probably not be an embedding irrespective of $f$. 
Take for example the case where $K$ is the following four-vertex graph embedded in $\mb R^m=\mb R^2$
\[\xymatrix{
&&& \bullet\,a\\
&&& *++[o][F-]{v_1}\\
&&& *++[o][F-]{v_2}\ar@{-}[u]\ar@{-}[dll]\ar@{-}[drr]\\
\bullet\,b & *++[o][F-]{v_3}\ar@{-}[rrrr] &&&& *++[o][F-]{v_4} & \bullet\,c\\
&&& \bullet\,y 
}\]
and $\mc S=\{a,b,c,y\}\subset\mb R^2$.
The nearest vertex to $a$, $b$, $c$, and $y$ is $v_1$, $v_3$, $v_4$, and $v_2$ respectively,
so one iteration of k-means on this data results in the edge $\{v_1,v_2\}$ intersecting the edge $\{v_3,v_4\}$.  
On the other hand, the nearest point to $y$ in $K$ lies on the edge $\{v_3,v_4\}$, 
so a single iteration of Algorithm~\ref{LA1} results in $K$ being stretched out towards $\mc S$ without violating its embedding.   
In fact, for non-pathological $K$, 
Algorithm~\ref{LA1} has a strong tendency to retain $g\wcolon\seqm{K}{}{\mb R^m}$ (and each $f^\ell$) as embeddings,
given that the initial fitting $f\wcolon\seqm{K}{}{\mb R^m}$ is an embedding.  

\subsection{Refinement} 
Once we have obtained a sufficiently good fitting $g\colon\seqm{K}{}{\mb R^m}$ from Algorithm~\ref{LA1},
the next step is to refine it by getting rid of redundant simplices. These are the simplices $\sigma\subseteq K$ 
whose linearly embedded image $g(\sigma)$ has no points $g(y')\in g(\sigma)$ positioned near its interior, but instead, near its boundary $\bd g(\sigma)=g(\bd\sigma)$. 
We then project those $g(y')$ near the boundary into one of the simplices on the boundary in order to obtain a lower dimensional approximate representation of $y$. 
This process is repeated for the projections.

\begin{algorithm}[H]
\caption{Second stage: deleting redundant simplices, reducing dimension}
\label{LA2}
\begin{algorithmic}[1]
\STATE \textbf{input:} a threshold value $\alpha\geq 0$, and each $y'$, $\sigma_y$, and $g\colon\seqm{K}{}{\mb R^m}$ from Algorithm~\ref{LA1};
\FOR{each $y\in \mc S$} 
\STATE set $z\leftarrow y'$ and $\sigma\leftarrow \sigma_y$; 
\WHILE{$\sigma$ is not a vertex}
\STATE find a choice of $\tilde z$ in $\bd\sigma$ such that $g(z)$ is nearest to $g(\tilde z)$,
and $\tilde\sigma\subseteq\bd\sigma$ the smallest dimensional boundary simplex that contains $\tilde z$;
\IF{$||g(\tilde z)-g(z)||\leq \alpha$}
\STATE set $z\leftarrow\tilde z$ and $\sigma\leftarrow\tilde\sigma$; 
\ELSE
\STATE exit the while loop;
\ENDIF
\ENDWHILE
\STATE set $\tilde y\leftarrow z$ and $\tilde\sigma_y\leftarrow\sigma$, and $\tilde\lambda_{jy}$ (for each $v_j\in\tilde\sigma_y$)
the barycentric coordinates of $\tilde y$ in $\tilde\sigma_y$;
\ENDFOR
\STATE set $\tilde K\leftarrow \bigcup_{y\in\mc S}\tilde\sigma_y$;
\STATE \textbf{return} $\tilde K$, and each $\tilde y$, $\tilde\sigma_y$, and $\tilde\lambda_{jy}$.
\end{algorithmic}
\end{algorithm}

The output $\tilde K$ is a subcomplex of $K$ with $g\colon\seqm{K}{}{\mb R^m}$ restricting to a linear map 
$$
g\wcolon\seqm{\tilde K}{}{\mb R^m},
$$
and if $g$ is a good fitting, $g(\tilde K)$ is a locally linear approximation of $\mc S$,
and each $g(\tilde y)$ will be a close approximation of the corresponding $y\in\mc S$.
 The distance threshold $\alpha$ is fixed for simplicity. 
A better choice would be to have it monotonically decrease in each iteration to counteract the effect of the distance between $y'$ and $g(z)$ increasing in each iteration. 
Alternatively, a scale independent measure that is more tolerant of noise in larger scale structures might be desirable. 
This can be obtained for example by taking the minimum of the barycentric coordinates of $y'$ in $\sigma$ as a measure of nearness to the boundary.
Algorithm~\ref{LA2} also always keeps those simplices that have at least one point $y'$ near its interior with respect to $\alpha$,
though we could opt for something less sensitive to compromise accuracy for simplicity.

Depending on how many simplices are deleted from $K$ to get to $\tilde K$, and how far down we go in projecting each $g(y')$ into boundary simplices, 
$\tilde K$ will typically have lower dimensional facets than $K$. At the same time, its facets and simplices typically have lower dimension than $\mb R^m$, 
so each $y$ will have a lower dimensional approximate representation in terms of the barycentric coordinates $\tilde\lambda_{jy}$ of $\tilde y$ in 
$\tilde\sigma_y$. 
For example, if $\tilde\sigma_y$ happens to be a $1$-dimensional simplex (edge) with vertices
$v_{j_1}$ and $v_{j_2}$, then $y\approx g(\tilde y)=tg(v_{j_1})+(1-t)g(v_{j_2})$ where $(t,1-t)\in \mb R^2$ are the barycentric coordinates of 
$\tilde y$ in $\tilde\sigma_y$.
$y$ can then be approximately represented by the barycentric coordinates $(t,1-t)$ together with an assignment of simplex $y\mapsto\tilde\sigma_y$ 
(i.e. an assignment of those vertices in $\tilde\sigma_y$). In summary, points $y\in\mc S$ are assigned to linear combinations 
$$
\sum_{v_j\in \sigma_y} \tilde\lambda_{jy} g(v_j)
$$ 
where $\tilde\lambda_{jy}$ are the barycentric coordinates of of $\tilde y$ in $\tilde\sigma_y$.
This often gives a much more compact representation of $\mc S$, 
especially when $m$ is large and large patches of points in $\mc S$ lie on approximately linear subspaces of considerably lower dimension than $m$. 

\subsection{Some comparisons}
This approach to dimension reduction is similar to that taken by k-means and self-organizing maps (SOM)~\cite{Kohonen1982,Kohonen:2001:SM:558021}. 
The assignment of vertices (\emph{nodes}, \emph{neurons}, \emph{classes}, or \emph{clusters}, in different language) however does not have to be discrete, 
so the reduction is not necessarily one to dimension $0$ for every $y\in\mc S$. 
Instead, it can be seen as a form of Fuzzy k-means~\cite{Bezdek:1981:PRF:539444,YANG19931} -- 
in this case, multiple vertices in $g(\tilde K)$ are assigned to $y$ to varying degrees, 
this depending on which simplex is nearest to $y$ and where (in terms of barycentric coordinates) the nearest point $y'$ lies inside this simplex. 
Like SOM, the mapping $g$ is topology preserving, with adjacent vertices of $K$ and $\tilde K$ tending to have nearby points in $\mc S$ assigned to them.
SOM however acts directly on individual vertices by matching data points with nearest vertices instead simplices, 
and uses an explicit neighbourhood function on nearby vertices to effect this property.  
Since simplicial complexes are able to more efficiently reflect the shape of an object such as our point cloud $\mc S$,
one would expect that Algorithms~\ref{LA1} and~\ref{LA2} give a better approximation of $\mc S$ than classical SOM, at the same time using fewer vertices.
This is all at the expense of our dimension reduction generally being above dimension $0$ depending on $\tilde K$.

The end result of Algorithms~\ref{LA1} and~\ref{LA2} is that local patches in $\mc S$ are approximated by the affine subspaces of $\mb R^m$ determined by the simplices of $g(\tilde K)$.
This is similar to Cluster-PCA~\cite{Fukunaga:1971:AFI:1309278.1309587}, as well as the first stage of Locally Linear Embedding (LLE)~\cite{Saul:2003:TGF:945365.945372}.
In Cluster-PCA a collection of affine $d$-dimensional subspaces of $\mb R^m$ are iteratively fitted to their nearest points in $\mc S$ via PCA, 
while LLE uses convex combinations of $k$ nearest neighbours to each point in $\mc S$ (sampled from a smoothly embedded $d$-manifold) 
in order to obtain an approximation of a local coordinate patch. 
In our case $\mc S$ need not be a manifold, nor does the dimension $d$ of the affine subspaces have to be known beforehand. 
Instead, $K$ has to be selected to have a high enough dimension and enough redundant facets and vertices (but not so many so as to over-fit)
in order to guarantee a good approximation of $\mc S$. 
None-the-less, we are left with a convenient approximation of the shape of $\mc S$ in terms of $\tilde K$ and $g$.
These dimension reduction techniques aside, 
there are various shape approximation techniques known to the computer graphics community, 
where, for example, a mesh is iteratively fitted to data by solving certain least-squares optimization problems~\cite{Hoppe:1993:MO:166117.166119}.

\subsection{Further dimension reduction}

If $g\colon\seqm{\tilde K}{}{\mb R^m}$ happens to be an embedding and $\tilde K$ has dimension $d$ with $d$ much smaller than $m$,
then one would like to go a step further and obtain a lower dimensional embedding $h\wcolon\seqm{\tilde K}{}{\mb R^k}$ for some $d\leq k<m$
that preserves as much of the properties of $g$ as possible (preserving geodesic distance for instance when $\tilde K$ is a manifold). 
Assuming our fitting $g$ is good, a dimension reduction of $\mc S$ into $\mb R^k$ follows since each $\tilde y\approx y\in\mc S$ and $\tilde y$ is in $g(\tilde K)$. 

There is a history of theoretical work dealing with the question whether \emph{any} embedding into $\mb R^k$ 
of a $d$-dimensional simplicial complex $\tilde K$ can exist in the first place (ignoring $g$ for the time-being). 
One of the most well-known results is Van Kampen's generalization of Kuratowski's graph planarity condition~\cite{Kuratowski1930,MR3069580,MR0089410,MR0215305}, 
which gives an obstruction to embedding $\tilde K$ into $\mb R^{2d}$ lying in the degree $2d$ cohomology of a certain \emph{deleted product} subcomplex of $\tilde K\times\tilde K$.
In fact, this is the largest dimension of interest since any simplicial complex of dimension $d$ can be embedded linearly into $\mb R^{2d+1}$ 
(for instance, by mapping each vertex to a unique point on the parametric curve $(t,t^2,\ldots,t^{2d+1})$). 
Some more recent work has focused on PL-embedding into $\mb R^{d+1}$~\cite{2016arXiv160501240B}, 
or the tractability of embedding as a decision problem~\cite{DBLP:journals/corr/abs-0807-0336}.

A direct computational approach is possible when $g(\tilde K)$ is a triangulation of an embedded manifold; for example, 
by applying Isomap~\cite{TSL} to its vertices $g(v_j)$ and some choice of samples in $g(\tilde K)$.
Since the entire manifold $g(\tilde K)$ is already given, the geodesic distances between the samples can be given exactly, 
and what remains is the MDS stage of Isomap. In any case, $g(\tilde K)$ is not a sampled space, 
and moreover, it is often a much simpler object than $\mc S$ consisting of a much smaller collection of vertices and facets.  
Any dimension reduction of $\mc S$ by first reducing the dimension of the embedding of $\tilde K$ should be easier 
compared to a more head-on approach.

\subsection{Complexity}
The computationally intensive step that dominates Algorithm~\ref{LA1} involves finding the points $y'\in K$ for which $f^{\ell}(y')$ is nearest to $y\in\mc S$.
These points can be found as follows.

\begin{algorithm}[H]
\caption{Find nearest points on a simplex along with their barycentric coordinates}
\label{LA3}
\begin{algorithmic}[1]
\STATE \textbf{input:} $\mc S\subset\mb R^m$, a simplex $\sigma\subseteq K$, and linear map $f^\ell\colon\seqm{K}{}{\mb R^m}$; 
\IF{$\sigma$ is a vertex $v_j$}
\STATE set $\lambda^{\sigma}_{1y}\leftarrow 1$ and $y^\sigma\leftarrow f^\ell(v_j)$ for each $y\in\mc S$; 
\ELSE
\STATE let $r:=|\mc S|$ and pick some ordering of $\mc S=\{y_1,\ldots,y_r\}$; 
\STATE Let $M$ be the $m\times d$ matrix whose $i^{th}$ column is the vector $w_{i+1}-w_1$,
where $\{w_1,\ldots,w_{d+1}\}\subseteq \{f^\ell(v_1),\ldots,f^\ell(v_n)\}$ are the vertices of $f^\ell(\sigma)$;
\STATE Let $S$ be the $m\times r$ matrix whose $j^{th}$ column is the vector $y_j-w_1$;
\STATE compute the Moore-Penrose pseudoinverse $\bar M$ of $M$, the $d\times r$ matrix $B:=\bar M S$, 
the $1\times r$ matrix $B'$ whose $j^{th}$ entry is $1-\sum_i B_{ij}$,
and take the $(d+1)\times r$ matrix $\mc B:=\begin{bmatrix}B'\\B\end{bmatrix}$;
\STATE partition $\mc S$ into disjoint subsets $\bar{\mc S}$ and $\mc S_1,\ldots,\mc S_{d+1}$ where 
$$
\mc S_k := \cset{y_j\in\mc S}{(\mc B)_{kj}<0}-\cunion{i<k}{}{\mc S_i}
$$
and
$$
\bar{\mc S} := \mc S-\cunion{i}{}{\mc S_i};
$$
\STATE for each $y_j\in\bar{\mc S}$, set $\lambda^{\sigma}_{iy_j}\leftarrow (\mc B)_{ij}$ for $1\leq i\leq d+1$,
 and $y^\sigma_j\leftarrow\sum_i \lambda^{\sigma}_{iy_j}\omega_i$;
\FOR{each $\mc S_k$ that is non-empty} 
\STATE let $\sigma_k$ be the boundary simplex of $\sigma$ containing every vertex $w_i$ except $w_k$,
and set $\lambda^{\sigma}_{ky_j}\leftarrow 0$;
\STATE repeat this algorithm for $\sigma_k$ in place of $\sigma$ and $\mc S_k$ in place of $\mc S$,
letting $y^\sigma:=y^{\sigma_k}$ and $\lambda^{\sigma}_{iy}:=\lambda^{\sigma_k}_{iy}$ (for $i\neq k$) denote its output for each $y\in\mc S_k$;
\ENDFOR
\ENDIF
\STATE \textbf{output:} $y^\sigma$ and $\lambda^{\sigma}_{iy}$ for each $y\in\mc S$.
\end{algorithmic}
\end{algorithm}

The output $y^\sigma$ is the nearest point to $y$ lying on $f^\ell(\sigma)$, 
and the $\lambda^{\sigma}_{iy}$'s are its barycentric coordinates in $f^\ell(\sigma)$.
The smallest dimensional simplex $\sigma_y\subseteq\sigma$ for which $y^{\sigma}$ is in $f^\ell(\sigma_y)$ can quickly be found
by looking at which of the $\lambda^{\sigma}_{iy}$'s are zero.

The product $\bar M S$ and pseudoinverse $\bar M$ can be determined using singular value decomposition of $M$ in $\mc O(dmr)$ time and $\mc O(\min\{m^2d,md^2\})$ time respectively,
so all steps before the final for-loop can be executed in $\mc O(dmr+\min\{m^2d,md^2\})$. Typically $r\geq\max\{m,d\}$, in which case this simplifies to $\mc O(dmr)$.
Since the recursive call in each iteration of the for-loop is executed only for those dimension $d-1$ boundary simplices $\sigma_i\subseteq\sigma$ for which $\mc S_i$ is non-empty,
and since the $\mc S_i$'s are all disjoint subsets of the current input $\mc S$, then at most $r_d=\min\{r,(d+1)!\}$ recursive calls are made.
So the total execution time of the above algorithm is $\mc O(dmrr_d)$. 

The nearest point $f^\ell(y')$ to $y$ is given as a choice of $y^\sigma$ over all facets (not simplices) of $\sigma\subseteq K$ 
that minimizes $||y^{\sigma}-y||$ ($y'$ itself is determined by the $\lambda^{\sigma}_{iy}$'s).
As a result, the time complexity of each iteration of Algorithm~\ref{LA1} is at worst $\mc O(Ndmrr_d)$ when $r\geq\max\{m,d\}$, 
where $N$ is the number of facets of $K$ and $d$ is the dimension of the highest dimensional facet in $K$.  
This is largely due to the cost of finding each $y'$ and $y^\sigma$ precisely. 
Significant improvements can be made if an approximation scheme is used. 
For example, since $f^\ell$ does not change much in consecutive iterations in Algorithm~\ref{LA1} for large enough $\ell$,
$y'$ can be approximated in the $(\ell+1)^{th}$ iteration by searching only those facets adjacent to the simplex that contains $y'$
in the previous iteration. In effect, this means sending an appropriate subset $\mc S_\sigma\subseteq\mc S$ in place of $\mc S$ 
as input for Algorithm~\ref{LA3}.
At the same time, Algorithm~\ref{LA1} is easy to parallelize, since Algorithm~\ref{LA3} can be applied simultaneously to each facet $\sigma$ of $K$.

\section{Some Examples on Synthetic Data}

Algorithm~\ref{LA1} was tested without the pruning stage (Algorithm~\ref{LA2}).
This was done on data $\mc S$ sampled from $1$, $2$, and $3$ dimensional spaces using grids, triangulated meshes, or their boundaries $K$ 
embedded into $\mb R^2$, $\mb R^3$, and $\mb R^4$. 
Precise values for nearest points were found using Algorithm~\ref{LA3}, without any form of approximation, while the learning rate $s$ was set at $0.1$.
Implementation is in MATLAB R2015b. Source code can be found at \url{https://github.com/pbebenSoton/smeans}.

\begin{figure}[H]
    \centering
    \begin{subfigure}[h]{0.3\textwidth}
        \includegraphics[width=\textwidth]{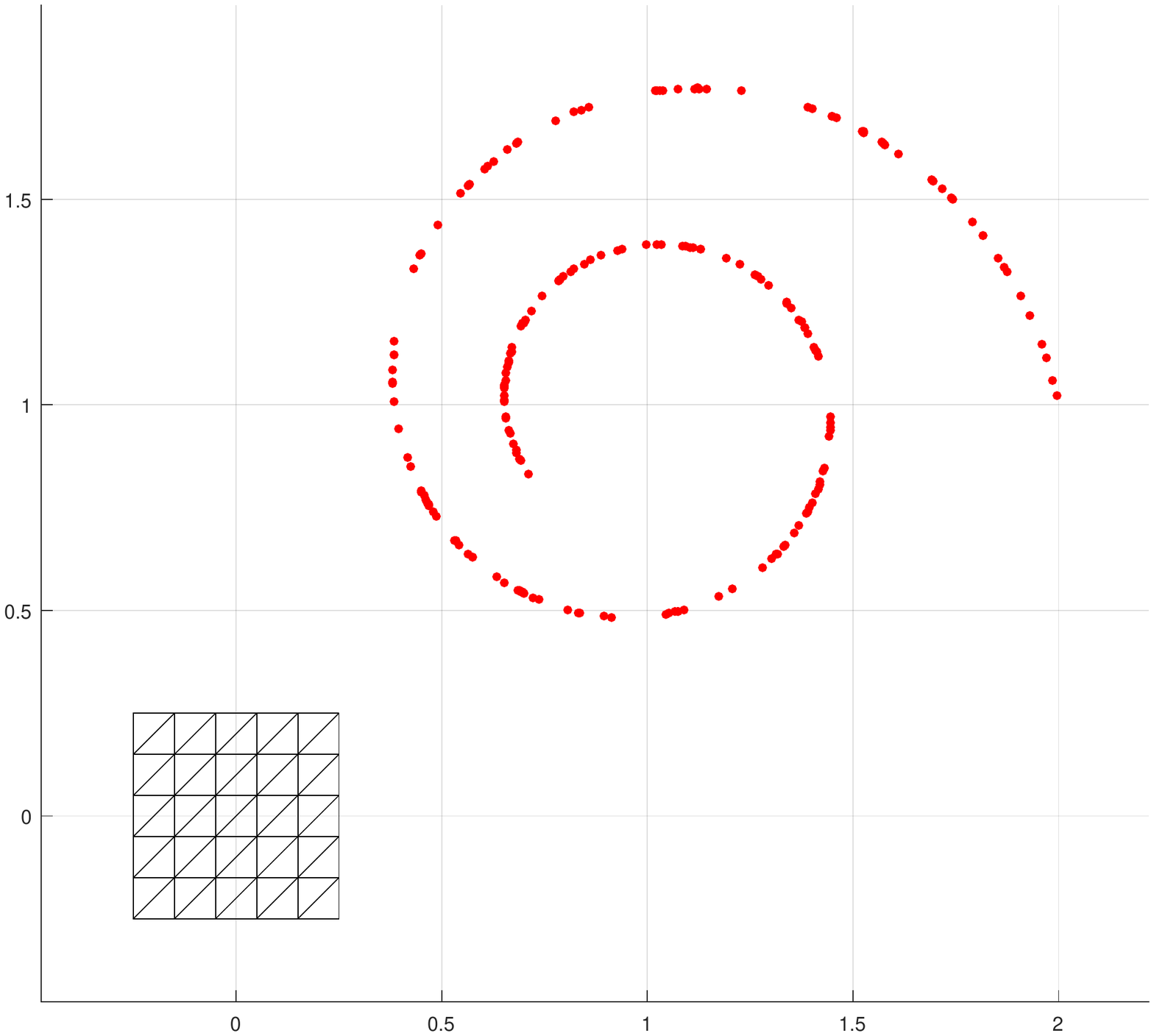}
        \caption{$0$ iterations}
    \end{subfigure}
    \begin{subfigure}[h]{0.3\textwidth}
        \includegraphics[width=\textwidth]{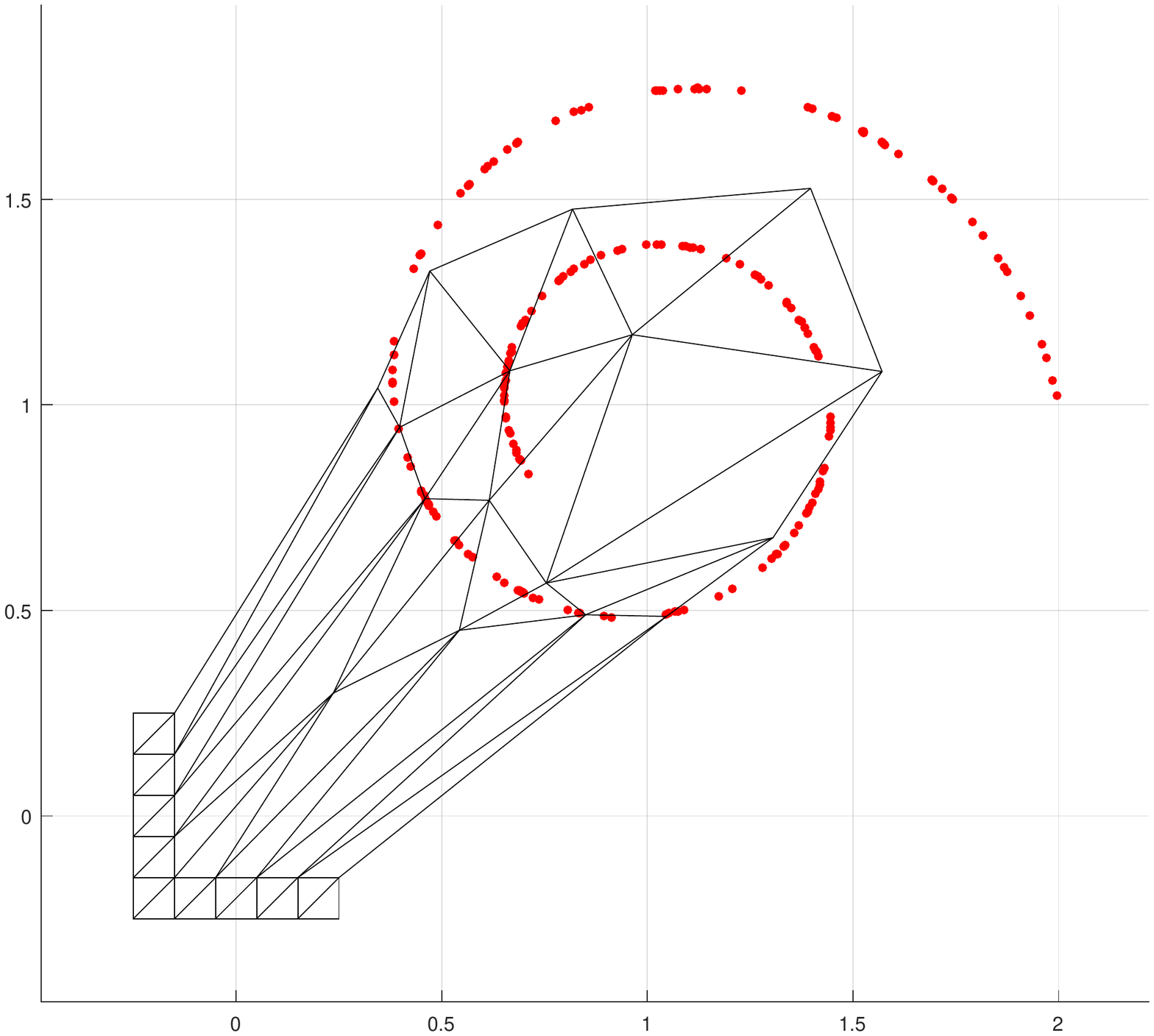}
        \caption{$20$ iterations}
    \end{subfigure}
    \begin{subfigure}[h]{0.3\textwidth}
        \includegraphics[width=\textwidth]{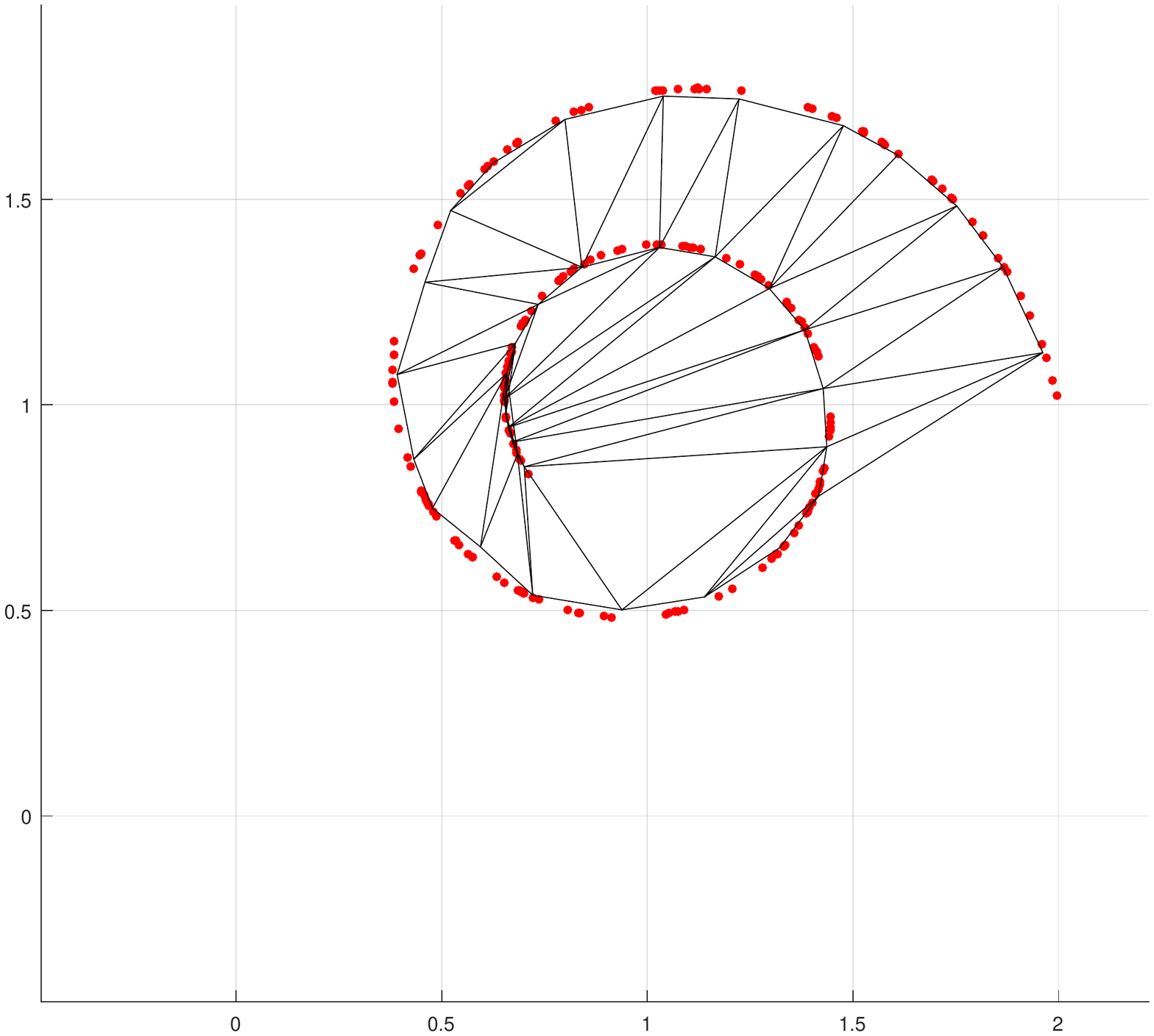}
        \caption{$150$ iterations}
    \end{subfigure}
    \caption{$200$ points sampled from a \emph{Swiss roll}, $K$ a $1$-dimensional $5\times 5$ grid. 
    The edges containing no points near their interior are deleted during the pruning stage.}\label{LF2}
\end{figure}

\begin{figure}[H]
    \centering
    \begin{subfigure}[h]{0.45\textwidth}
        \includegraphics[width=\textwidth]{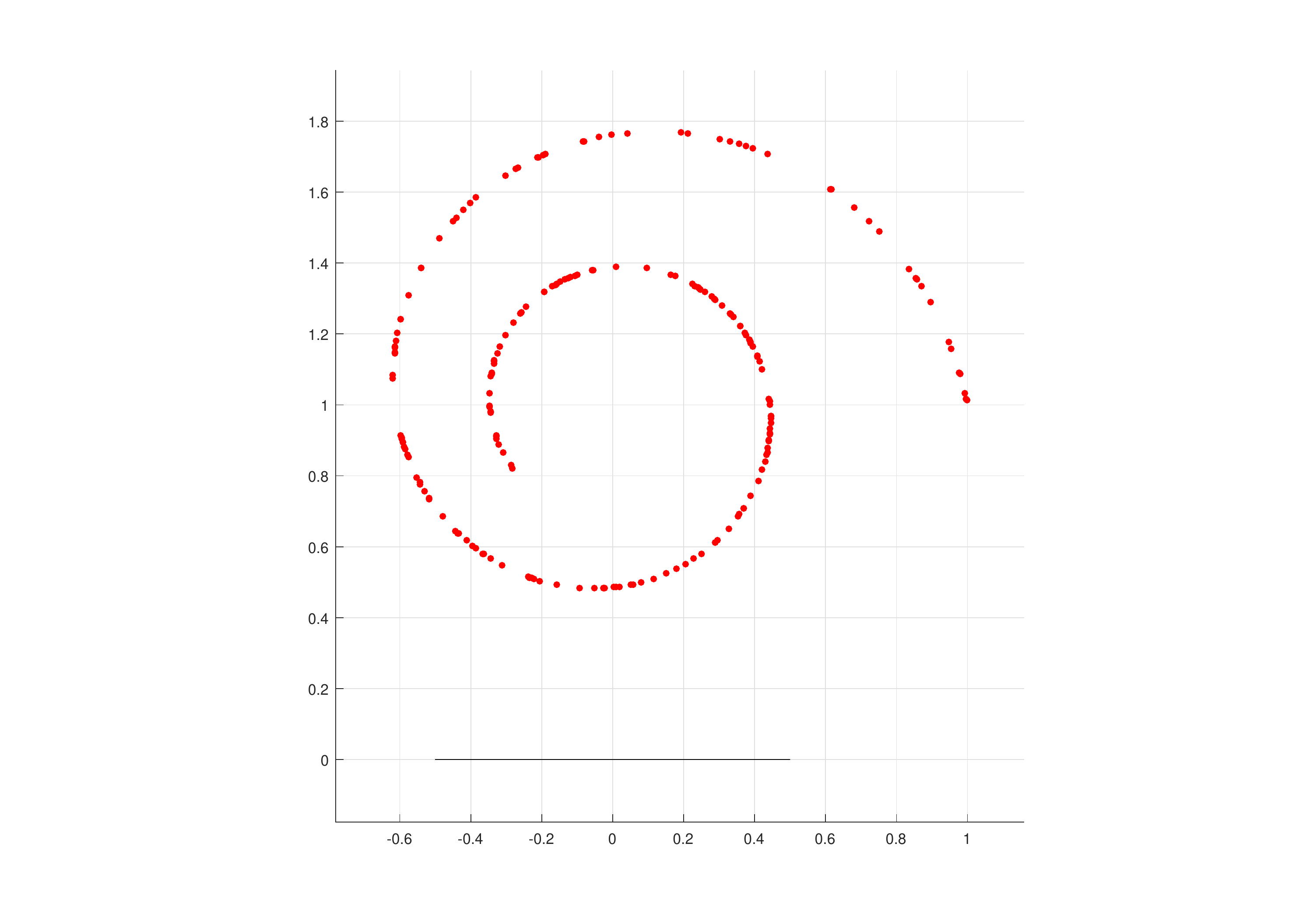}
        \caption{$0$ iterations}
    \end{subfigure}
    \begin{subfigure}[h]{0.45\textwidth}
        \includegraphics[width=\textwidth]{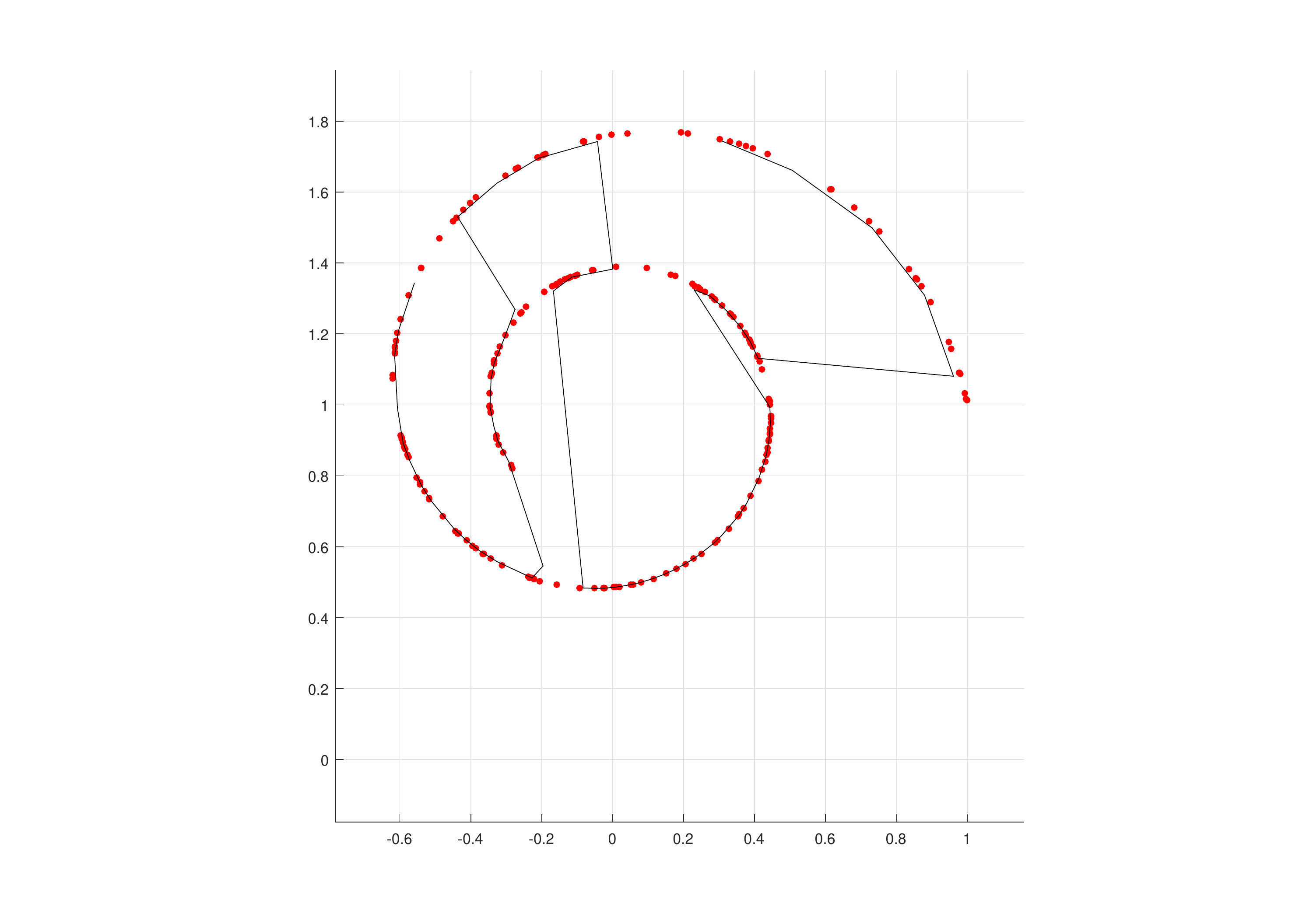}
        \caption{$200$ iterations}
    \end{subfigure}
    \caption{$200$ points sampled from a Swiss roll, $K$ a line subdivided into $60$ segments.
    The quality of this fitting is similar to classical SOM.
    Taking a grid of ambient dimension as in Figure~\ref{LF2} gives better results 
		with fewer vertices and a similar number of facets.}\label{LF1}
\end{figure}

\begin{figure}[H]
    \centering
    \begin{subfigure}[h]{0.3\textwidth}
        \includegraphics[width=\textwidth]{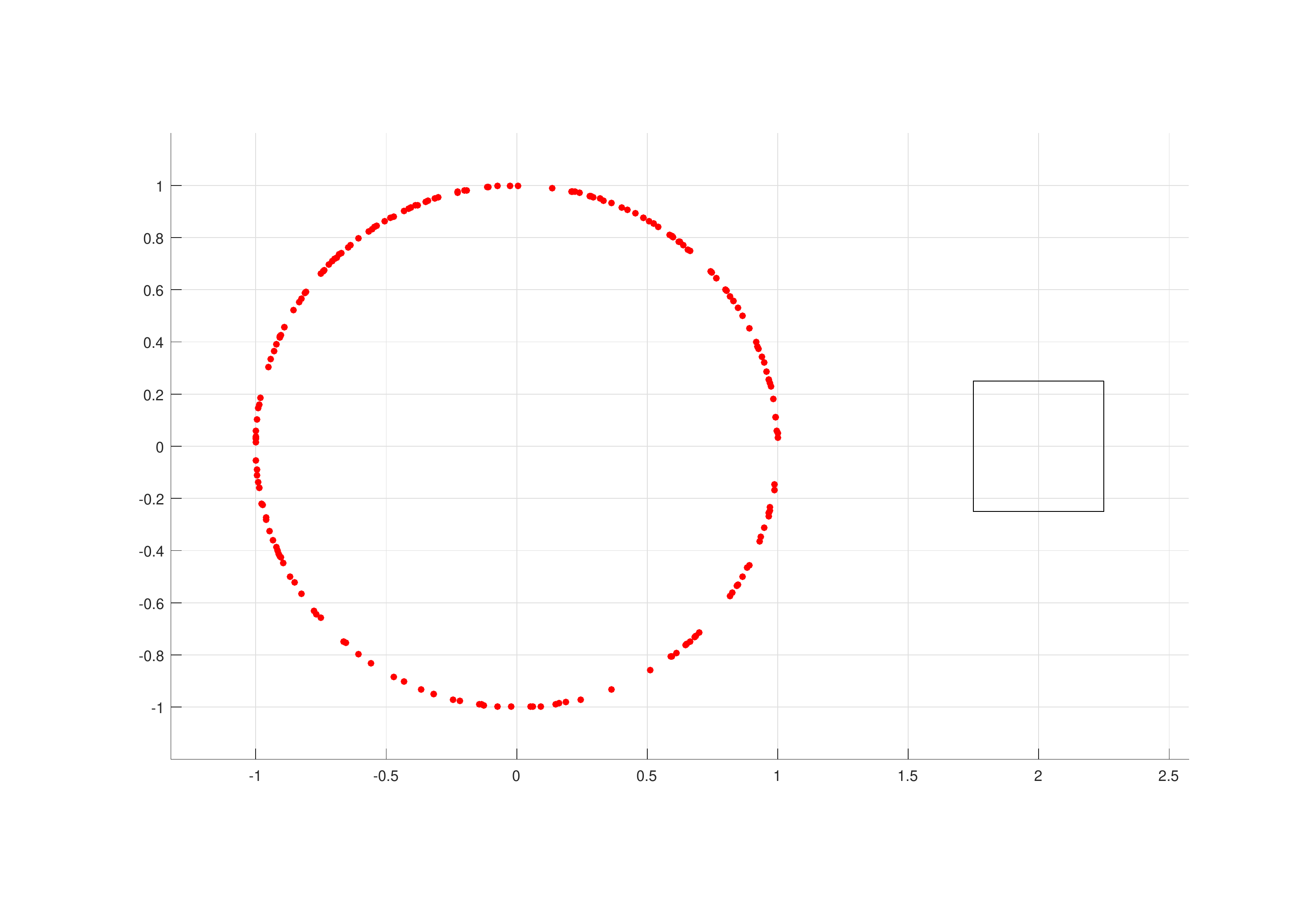}
        \caption{$0$ iterations}
    \end{subfigure}
    \begin{subfigure}[h]{0.3\textwidth}
        \includegraphics[width=\textwidth]{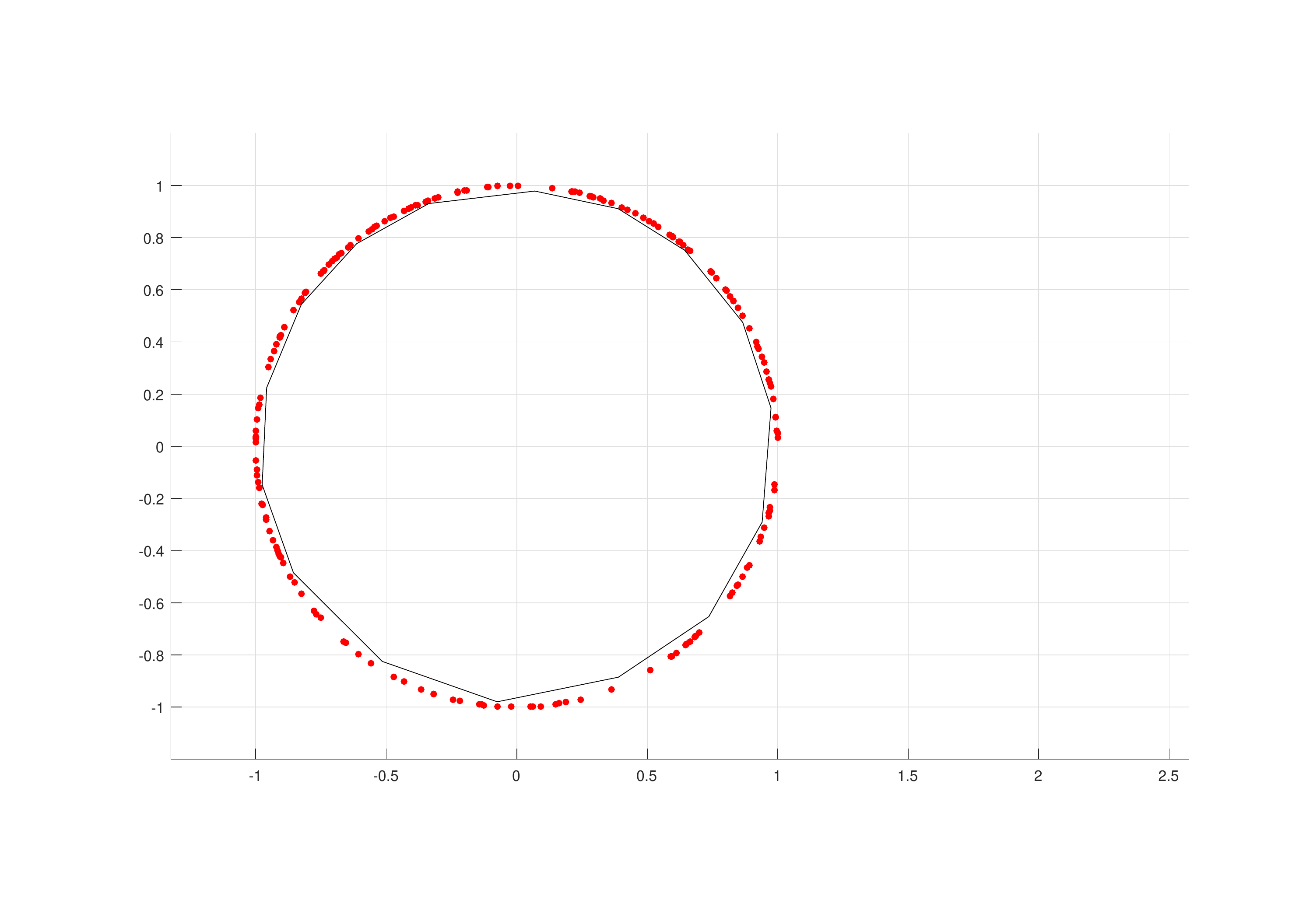}
        \caption{$150$ iterations}
    \end{subfigure}
    \caption{$200$ points sampled from a unit circle, $K$ a square with each side divided into $4$ segments.
		Fitted using the sets $\overline{\mc N}^\ell_j$ in place of $\mc N^\ell_j$.}\label{LF9}
\end{figure}

\begin{figure}[H]
    \centering
    \begin{subfigure}[h]{0.3\textwidth}
        \includegraphics[width=\textwidth]{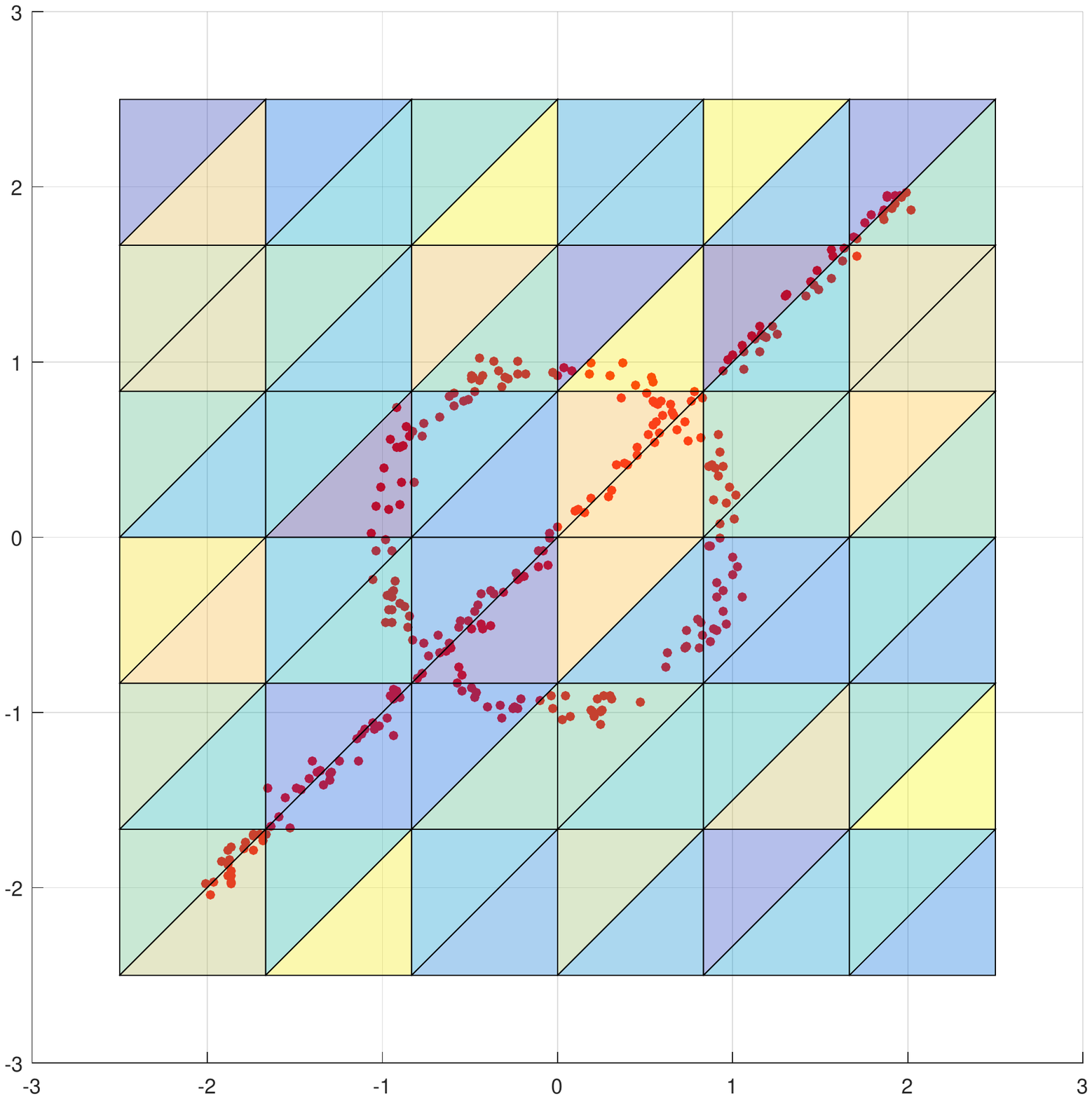}
        \caption{$0$ iterations}
    \end{subfigure}
    \begin{subfigure}[h]{0.3\textwidth}
        \includegraphics[width=\textwidth]{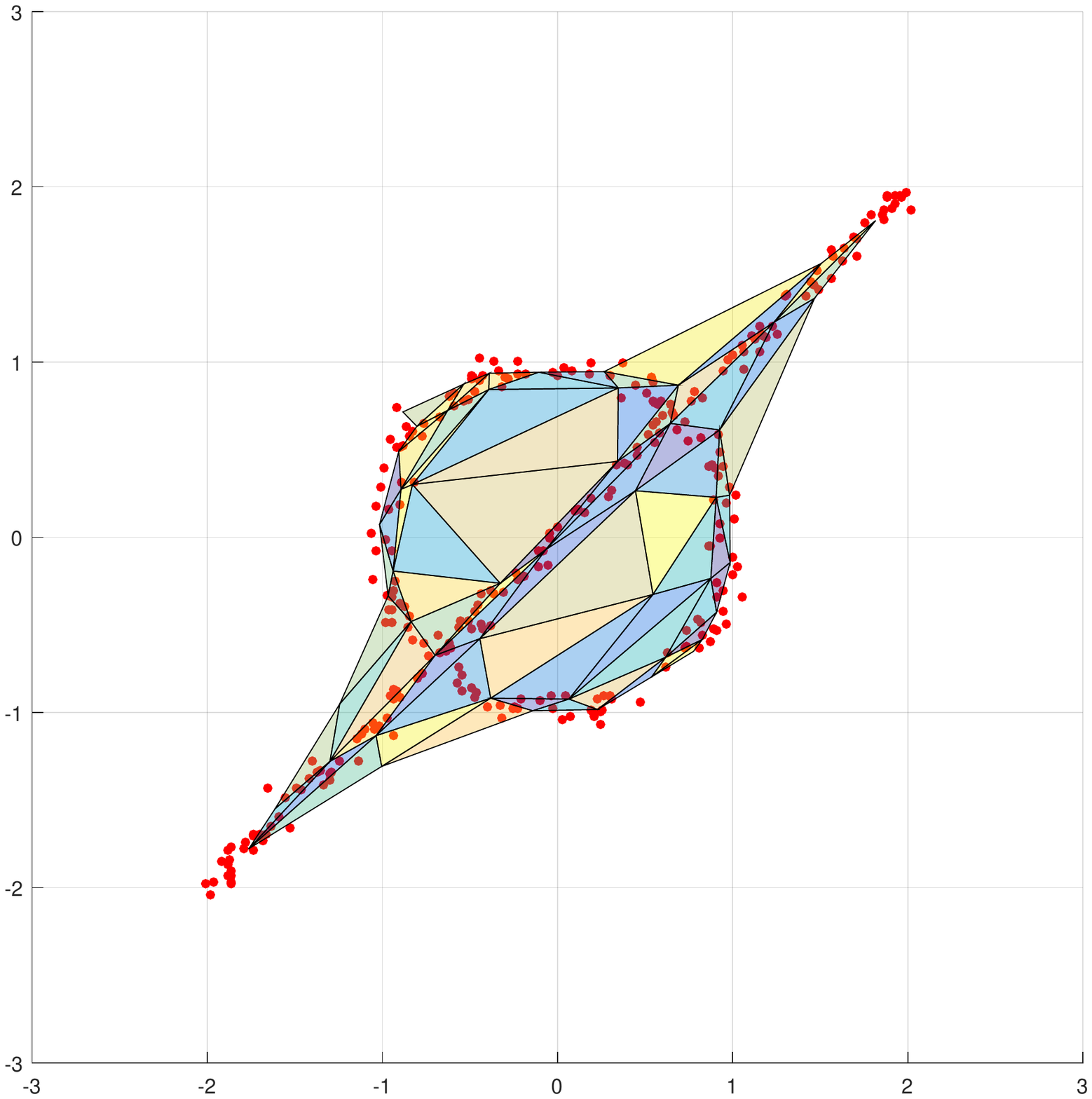}
        \caption{$100$ iterations}
    \end{subfigure}
    \caption{$300$ points sampled noisily from a circle and a line, $K$ a triangulated $6\times 6$ mesh.}\label{LF3}
\end{figure}

\begin{figure}[H]
    \centering
    \begin{subfigure}[h]{0.3\textwidth}
        \includegraphics[width=\textwidth]{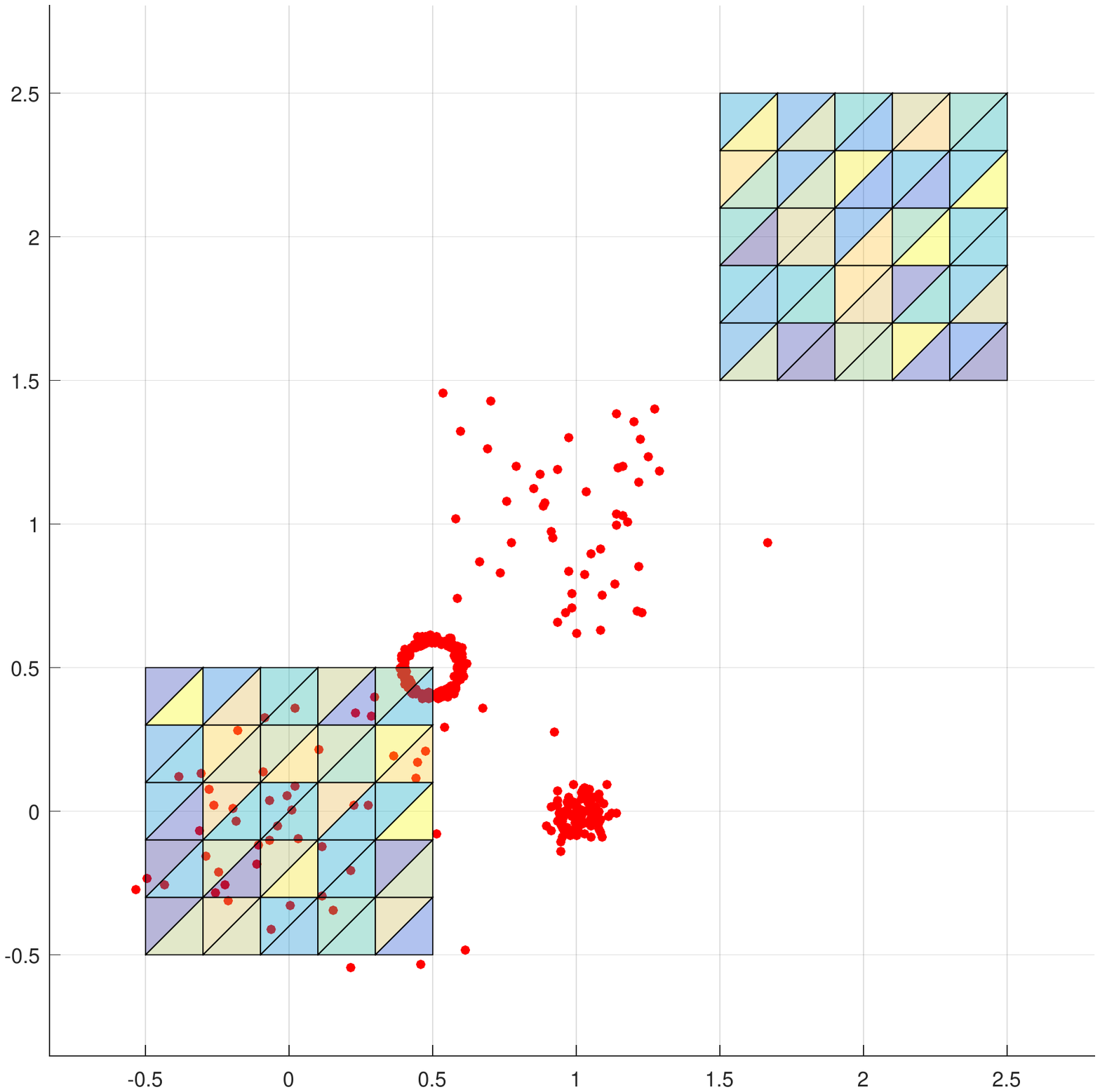}
        \caption{$0$ iterations}
    \end{subfigure}
    \begin{subfigure}[h]{0.3\textwidth}
        \includegraphics[width=\textwidth]{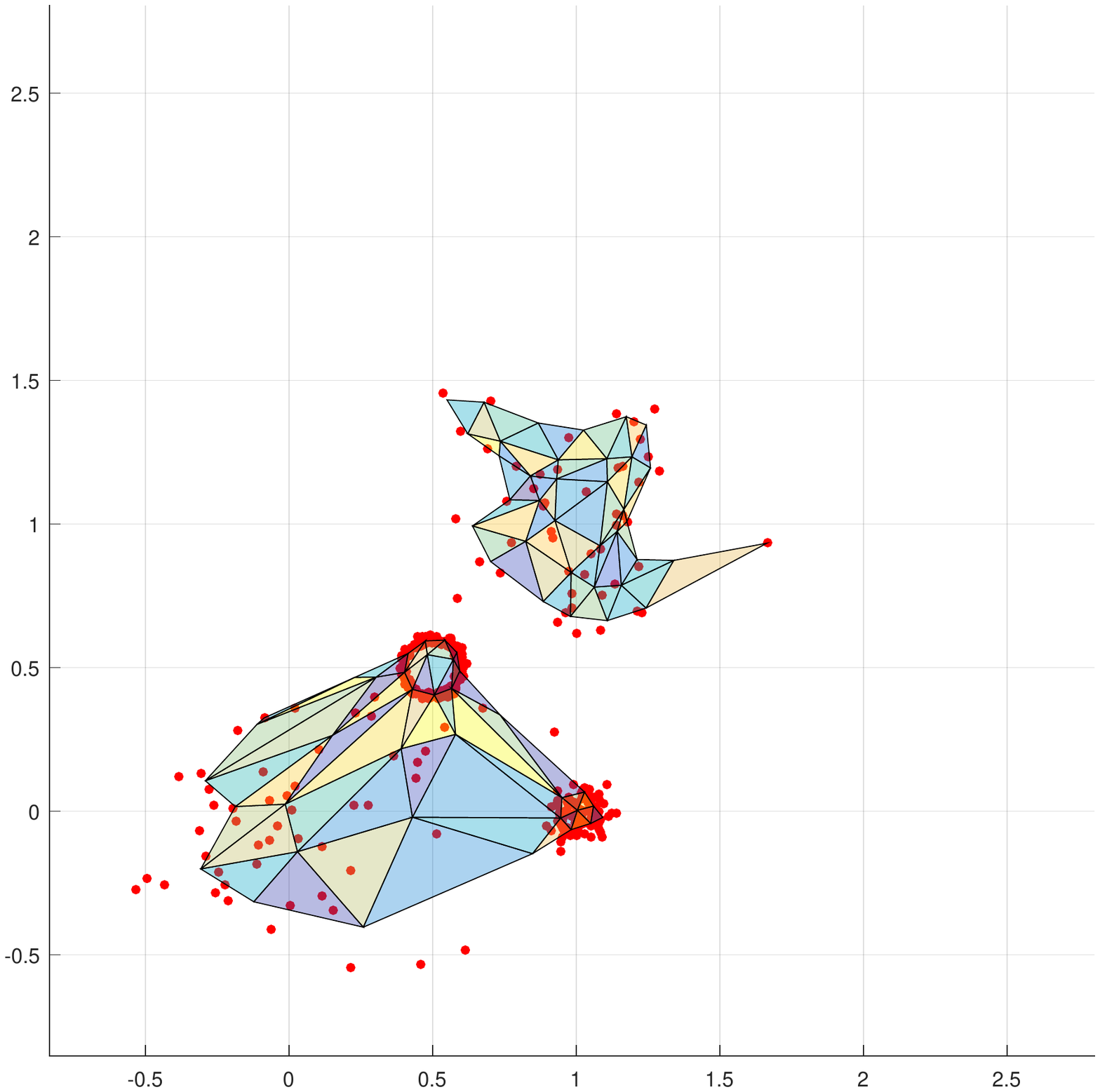}
        \caption{$150$ iterations}
    \end{subfigure}
    \caption{A noisy sampling of $350$ points containing a circle, 
    and $K$ a disjoint union of two $3\times 3$ mesh \emph{clusters}.
    Such clusters result in better fittings for highly disjoint data
    (compared to a single large mesh).}\label{LF6}
\end{figure}

\begin{figure}[H]
    \centering
    \begin{subfigure}[h]{0.3\textwidth}
        \includegraphics[width=\textwidth]{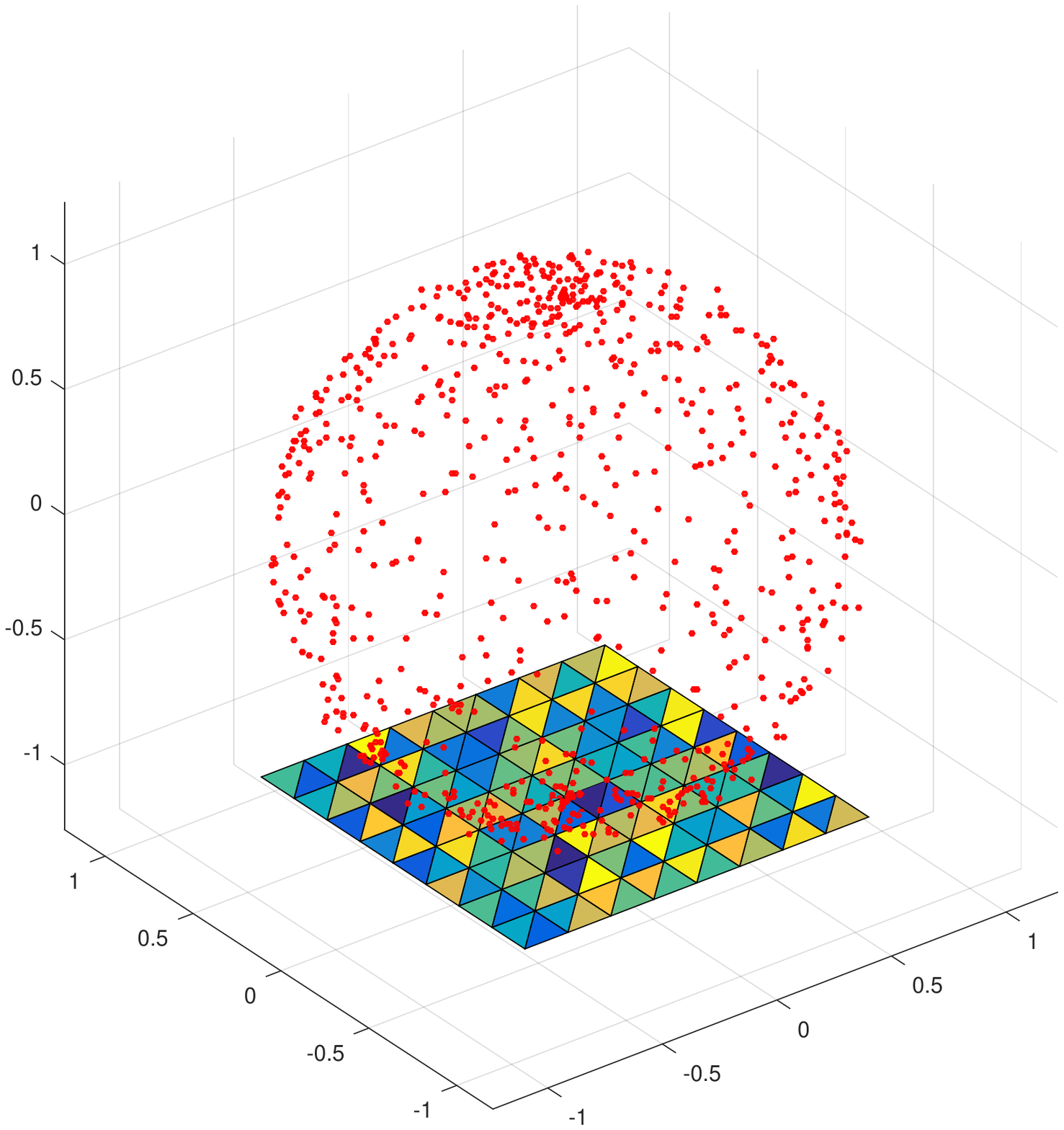}
        \caption{$0$ iterations}
    \end{subfigure}
    \begin{subfigure}[h]{0.3\textwidth}
        \includegraphics[width=\textwidth]{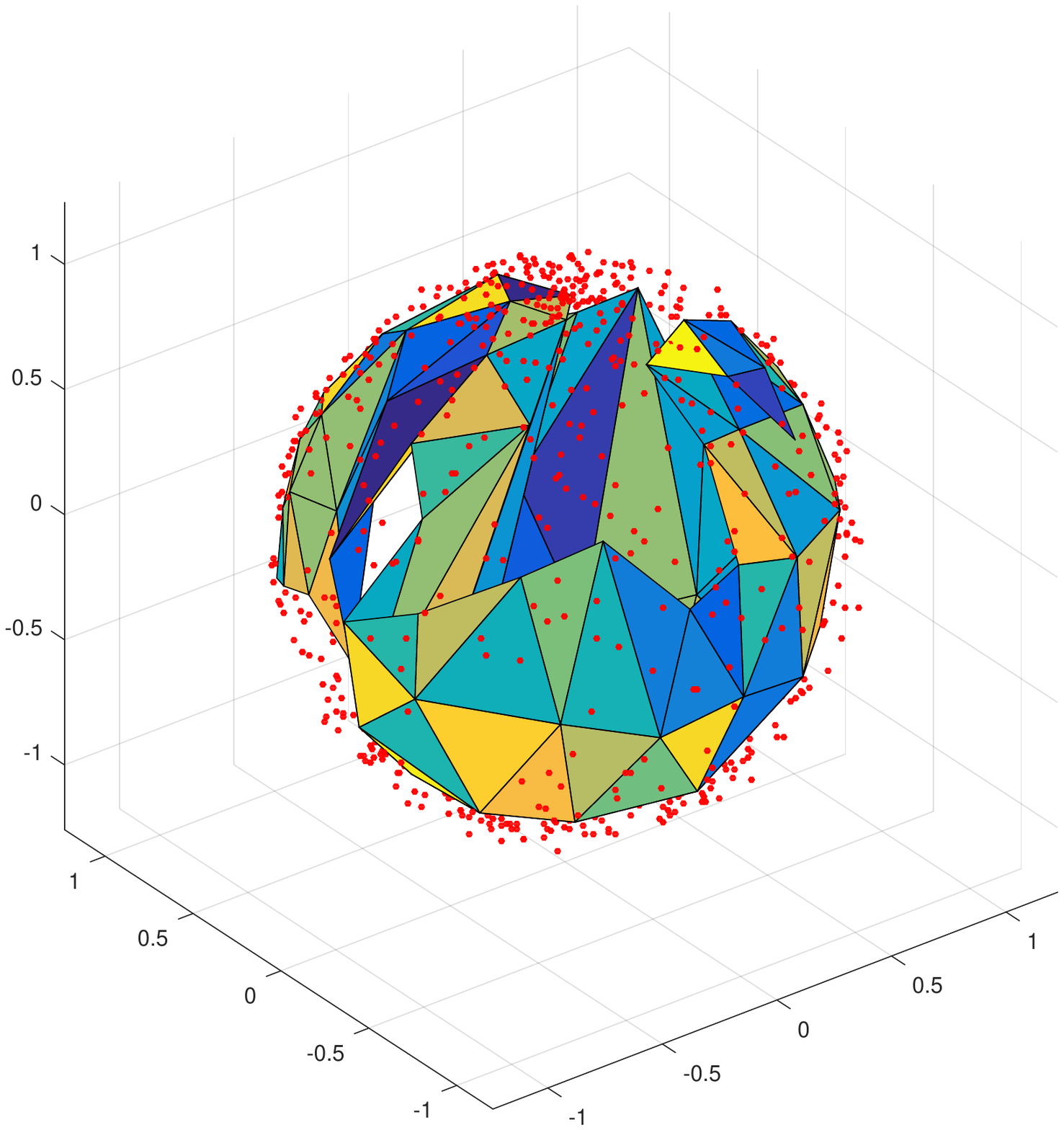}
        \caption{$150$ iterations}
    \end{subfigure}
    \caption{$800$ points noisily sampled from a unit $2$-sphere, $K$ a triangulated $8\times 8$ mesh.}\label{LF7}
\end{figure}

\begin{figure}[H]
    \centering
    \begin{subfigure}[h]{0.3\textwidth}
        \includegraphics[width=\textwidth]{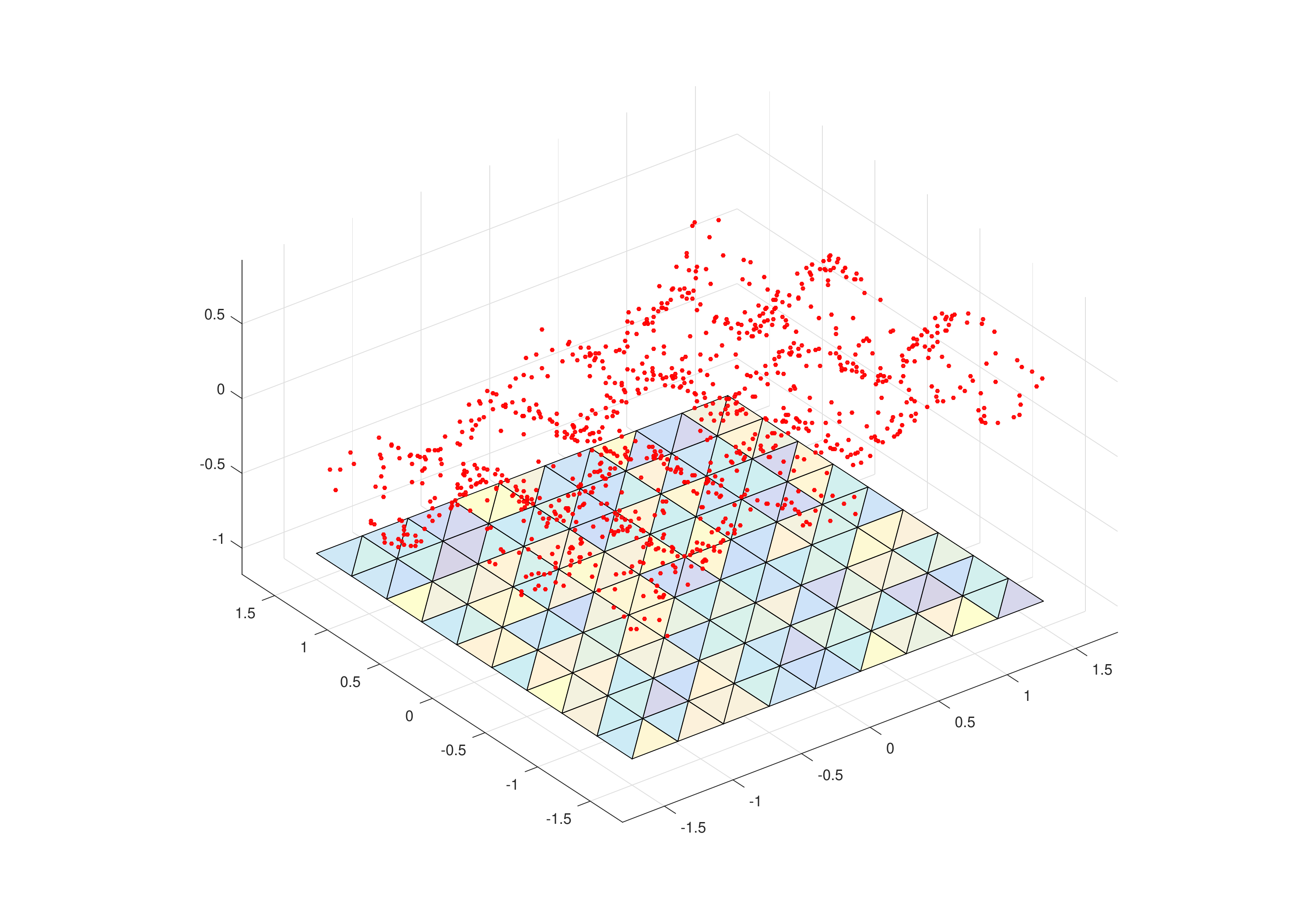}
        \caption{$0$ iterations}
    \end{subfigure}
    \begin{subfigure}[h]{0.3\textwidth}
        \includegraphics[width=\textwidth]{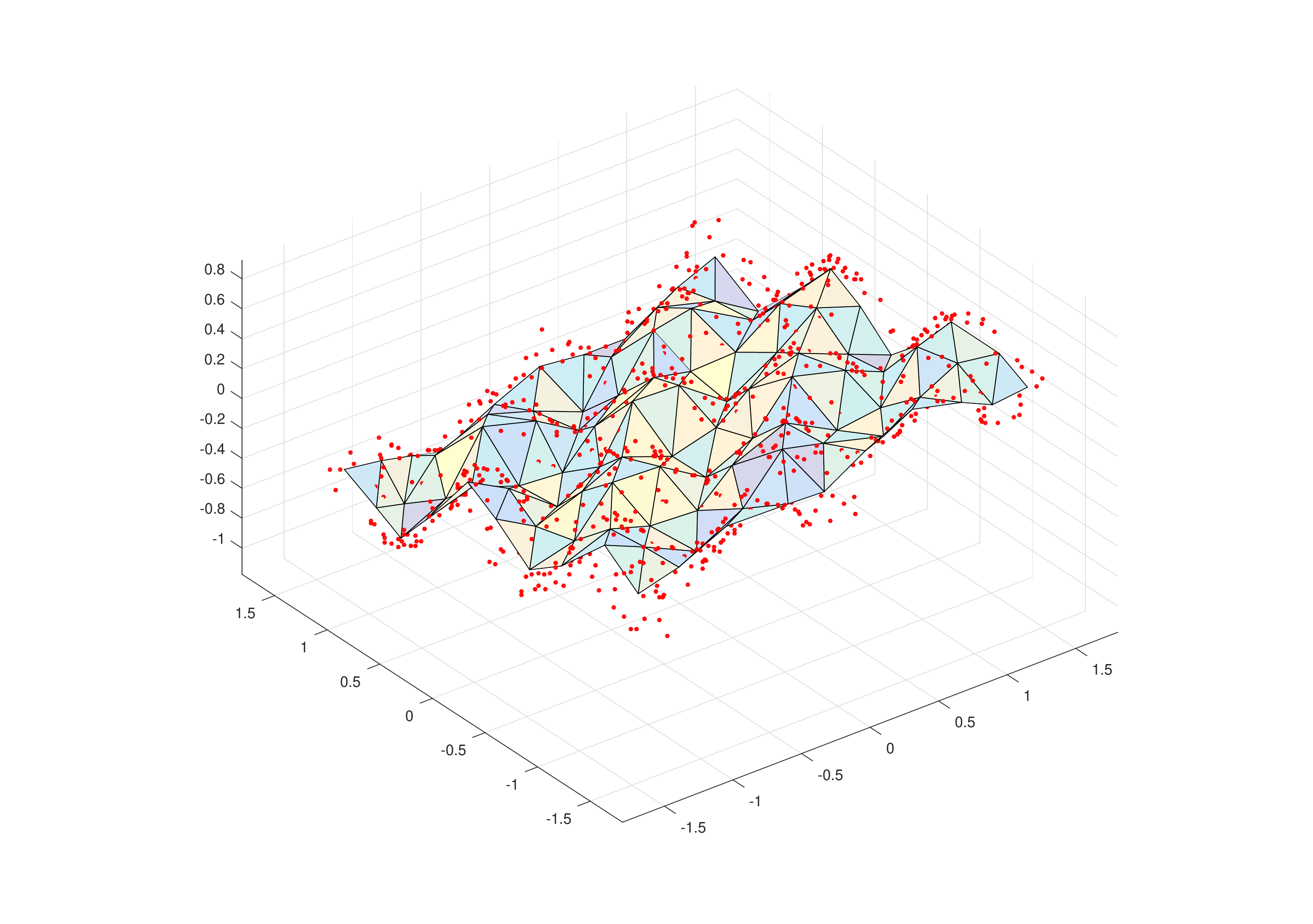}
        \caption{$40$ iterations}
    \end{subfigure}
    \caption{$1000$ points sampled from the surface $\frac{1}{3}cos(5x)sin(5y)+\frac{1}{5}(x-y)$, $K$ a triangulated $9\times 9$ mesh.}\label{LF4}
\end{figure}

\begin{figure}[h]
    \centering
    \begin{subfigure}[h]{0.3\textwidth}
        \includegraphics[width=\textwidth]{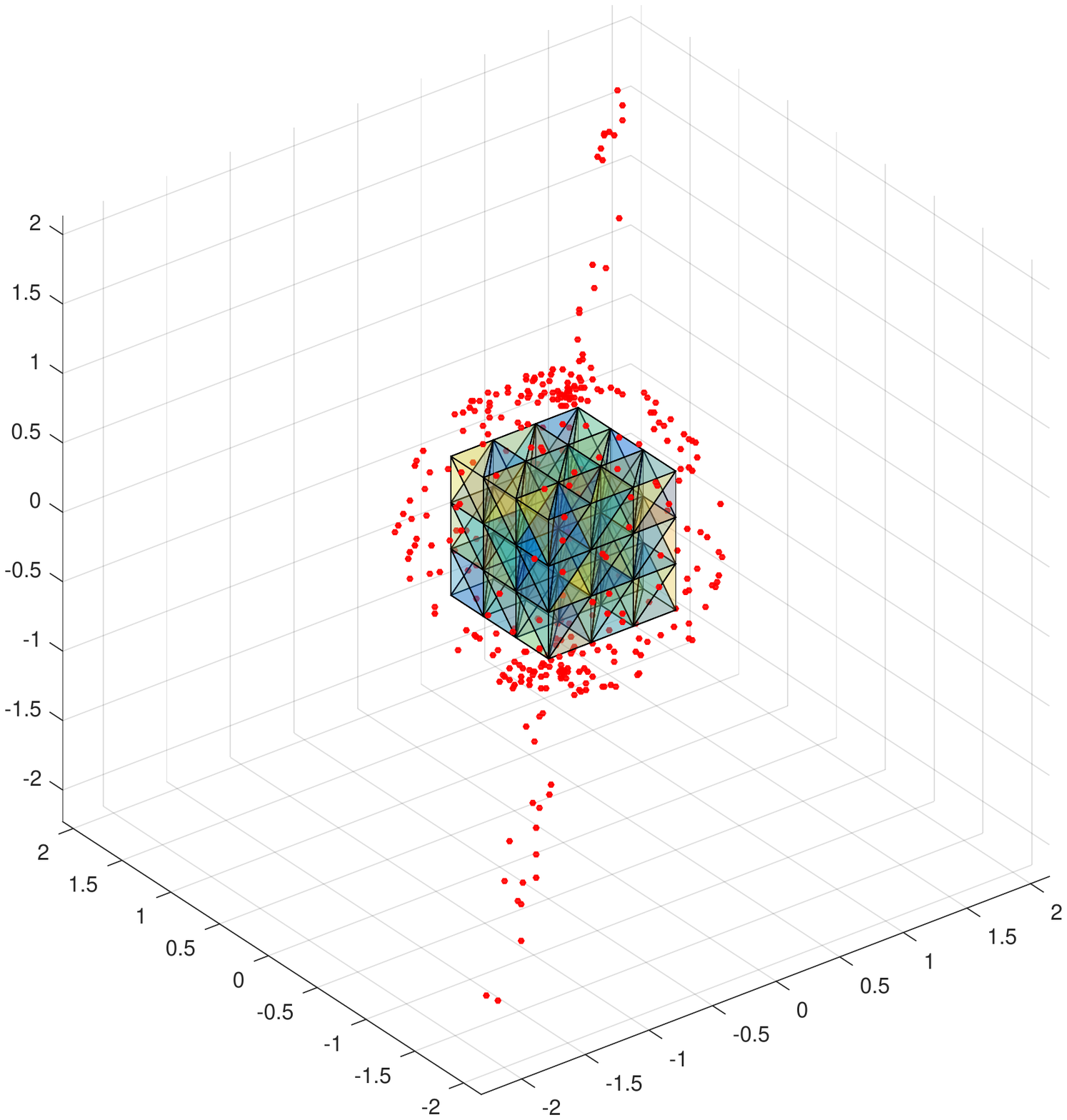}
        \caption{$0$ iterations}
    \end{subfigure}
    \begin{subfigure}[h]{0.3\textwidth}
        \includegraphics[width=\textwidth]{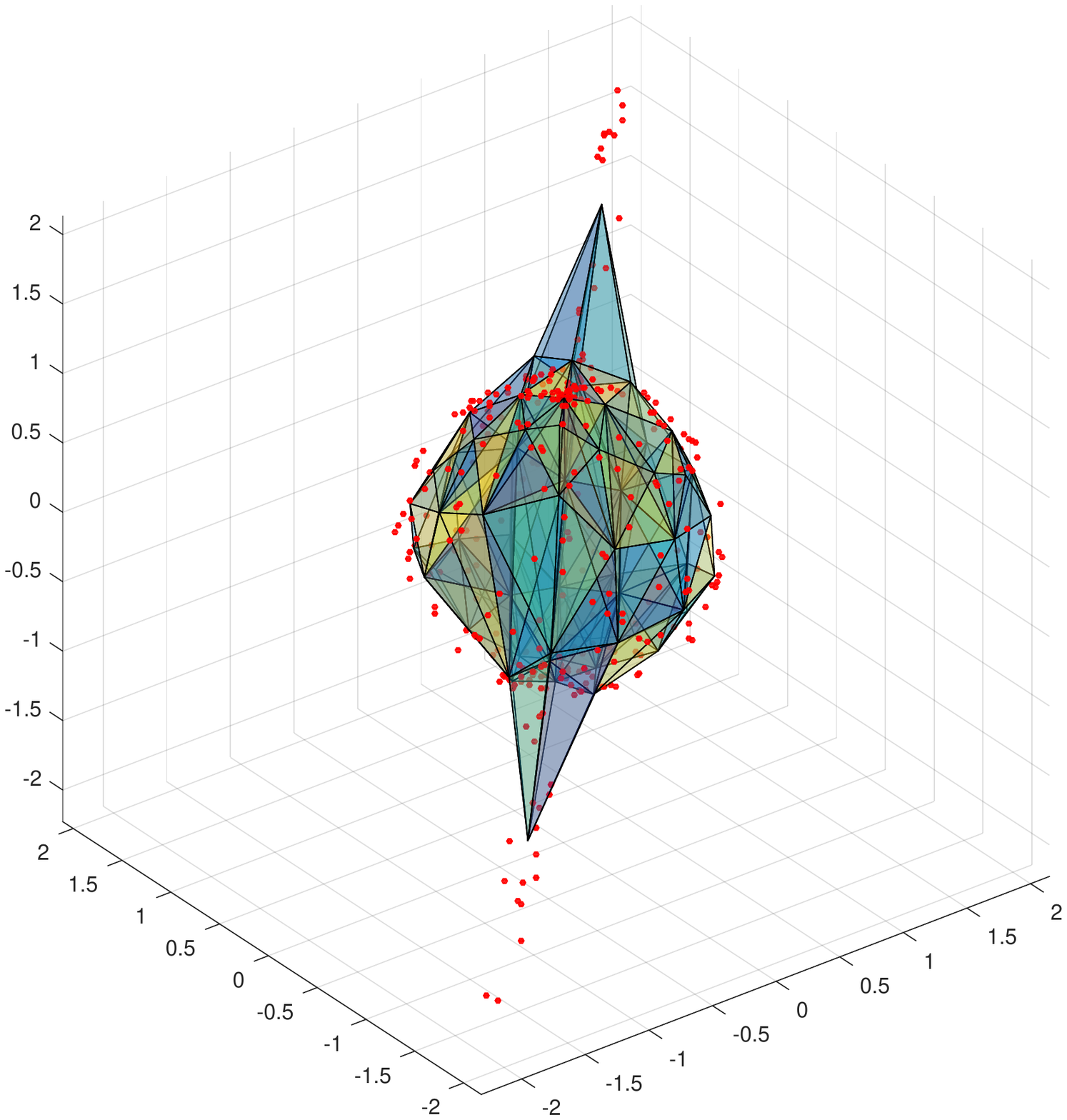}
        \caption{$40$ iterations}
    \end{subfigure}
    \caption{$300$ points noisily sampled from a $2$-sphere and a line, $K$ a triangulated $3\times 3\times 3$ mesh (solid cube).}\label{LF5}
\end{figure}

\begin{figure}[h]
    \centering
    \includegraphics[width=0.8\textwidth]{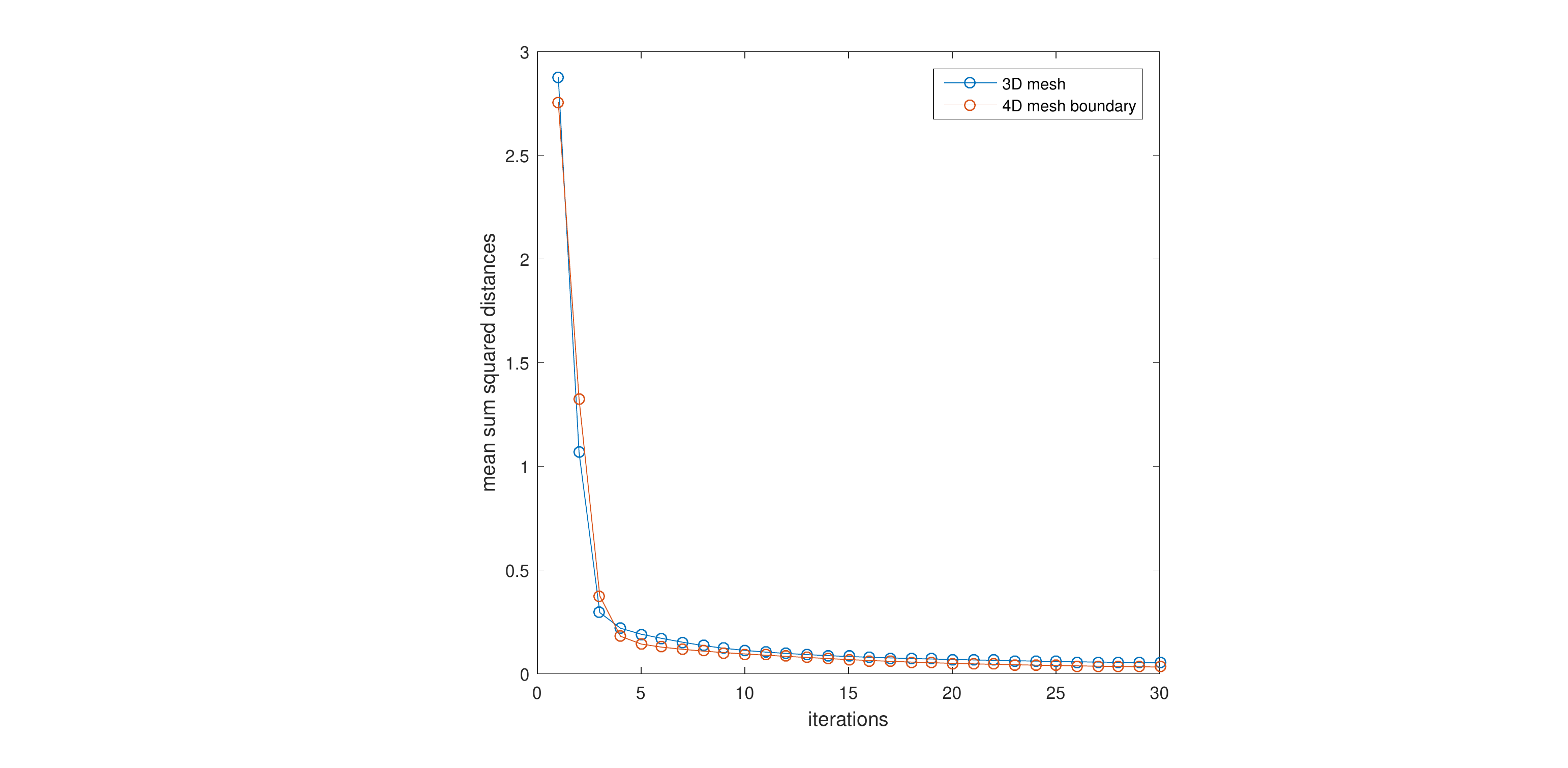}
    \caption{$\mc S$ taken as $2000$ points noisily sampled from a unit $3$-sphere embedded in $4$-dimensional euclidean space $\mb R^4$, 
		and $K$ respectively a triangulation of a $4\times 4\times 4$ mesh, and the boundary of a $3\times 3\times 3\times 3$ mesh.
		The mean sum of squared distances in each iteration is taken between points in $\mc S$ and their corresponding nearest points in the current fitting of $K$.}\label{LF10}
\end{figure}

\section{Discussion}

For non-pathological complexes $K$ (such as above) Algorithm~\ref{LA1} quickly converges to a stable fitting.
In most cases it also preserves the embedding of $K$, and where it does not, then it is usually very close to an embedding,
with only a few simplices intersecting over a relatively small area (intersections are more common for less noisy data).
All of these properties are less likely hold for irregularly triangulated $K$; for example, when the degree of some vertices is much higher than others.  

If there is convergence, what is the objective function that is being minimized? 
Since Algorithm~\ref{LA1} generalizes k-means clustering, an obvious first choice is the sum of squared distances 
$d_\ell=\sum_{y\in\mc S}{||y-f^\ell(y')||^2}$.
Indeed, there are cases where this function happens to be minimized between every successive iteration; 
for instance, in the example in Figures~\ref{LF4} and~\ref{LF10}.
On the other hand, if we start off as in Figure~\ref{LF3}, with $\ell=0$ and $f^0(K)$ containing every point in $\mc S$,  
then $f^\ell(K)$ is unlikely to contain $\mc S$ for large enough $\ell$, so $d_\ell$ will not always minimized. 
In this situation it appears more appropriate to consider the Hausdorff distance 
$d_H(\mc S,f^\ell(K))=\inf\cset{\epsilon\geq 0}{f^\ell(K)\subseteq (\mc S)_\epsilon\mbox{ and }\mc S\subseteq (f^\ell(K))_\epsilon}$ 
where $A_\epsilon$ is the $\epsilon$-neighbourhood of a subset $A\subseteq\mb R^m$.
In any case, it is not immediately clear what the objective function might be, or if there even is a single one.

Another question is: under what modifications to Algorithm~\ref{LA1} is preservation of embedding guaranteed for some given (or all) $K$?
One possibility is to assume all facets of $K$ have the same ambient dimension $m$ as $\mb R^m$, 
then to update the vertices $f^\ell(v_j)$ in each iteration one at a time --
at each update step considering only those $y\in\mc N^\ell_j$ that are inside a convex neighbourhood $B$ of $f^\ell(v_j)$, 
where $B$ is contained in the union of the (currently updated) facets which have $f^\ell(v_j)$ as a vertex.
This can be shown to preserve embedding, it keeps vertices on the boundary fixed, 
and therefore points $y$ that are inside $f^\ell(K)$ remain as such in subsequent iterations. 
Then in order to get a good fitting, a large enough triangulation of (say) an $m$-ball $K$ is selected, 
together with a sufficiently expansive initial embedding $f$ such that: there are enough interior facets in $K$ to approximate $\mc S$. 
The main downside to this approach is that $K$ would have to be infeasibly large when $m$ is large. 

As with k-means clustering, there is the issue of selecting the appropriate number of clusters (vertices) to start with. 
On top of this there are many other factors, such as the number, dimension, and the arrangement of facets that go into describing
our simplicial complex $K$. Ideally, $K$ would be taken to be a mesh of dimension $d$ with a fine enough subdivision, 
embedded into $\mb R^d$ so as to contain $\mc S$. Such choice $K$ represents a brute force approach,
where the guarantee of a good fitting is the result of a high density of redundant simplices spread out uniformly in $\mb R^d$.  
For large $d$ this is of course impractical -- even the smallest triangulation of the $15$-cube
has $2^{15}$ vertices and over ten million facets of dimension $15$, increasing exponentially with $d$~\cite{SMITH2000131}. 
On the other hand, a simplex of any dimension $d$ has only one facet and $d+1$ vertices, so there is plenty of room for compromise. 
Skirting this issue by selecting the mesh $K$ to have fewer subdivisions or a lower dimension can leave a large subset $\mc S'$ 
of points in $\mc S$ poorly approximated by the final fitting $g$ of $K$ (see Figures~\ref{LF7} and~\ref{LF5} for example).  
One way around this is to fit a new complex $K'$ to the subset $\mc S'$, 
but this runs the risk of $\mc S'$ being a poor sampling of the underlying space. So there might be nothing meaningful left to fit.    
Instead, we can consider a \emph{growing} procedure akin to growing SOM (GSOM)~\cite{AHS}:
(1) first by considering the set $\mc C$ of those simplices $\sigma\subseteq K$ for which $g(\sigma)$ is a poor approximation
of the points $y\in \mc S$ nearest to its interior (i.e. $\csum{y'\in int(\sigma)}{}{||y-g(y')||}$ is large);
(2) forming a new complex $K'$ from $K$ by attaching a $(k+1)$-dimensional simplex to each $k$ dimensional simplex in $\mc C$;
(3) extending $g$ to a linear map $g'\colon\seqm{K'}{}{\mb R^d}$; (4) then refitting $K'$ to $\mc S$ starting with $g'$ as our initial fitting. 
Simplices are the smallest objects that can be attached; a more robust alternative would be to form $K'$ from $K$ by attaching 
a triangulation of the product $L\times \Delta^1=\cset{(x,y)}{x\in L,\,y\in\Delta^1} $ to the subcomplex $L=\bigcup_{\sigma\in\mc C}\sigma$, 
where $\Delta^1$ is the $1$-simplex (an edge). 
In either approach we increase the dimension of $K$ locally in order to capture the spread of local patches of data in $\mc S$.
Then reapplying the pruning stage (Algorithm~\ref{LA2}) just as we did before captures the underlying structure.


\nocite{caytonalgorithms}
\nocite{2014arXiv1403.2877S}
\bibliographystyle{amsplain}
\bibliography{mybibliography}

\end{document}